\documentclass[10pt,twocolumn,letterpaper]{article}

\usepackage[pagenumbers]{cvpr}

\definecolor{cvprblue}{rgb}{0.21,0.49,0.74}
\usepackage[pagebackref,breaklinks,colorlinks,allcolors=cvprblue]{hyperref}
\usepackage{algorithm}
\usepackage{algorithmic}
\usepackage{microtype}
\usepackage[most,skins]{tcolorbox} 
\usepackage{caption}
\usepackage{comment}
\usepackage{multirow}
\usepackage{amsmath}
\usepackage{amsfonts}
\usepackage{adjustbox}
\usepackage{makecell}
\usepackage[table]{xcolor}
\usepackage{colortbl}
\usepackage{graphicx}
\colorlet{PastaYellow}{yellow!10}
\definecolor{GroupColor1}{RGB}{220,230,255}
\definecolor{GroupColor2}{RGB}{220,255,220}
\definecolor{GroupColor3}{RGB}{255,220,200}
\definecolor{GroupColor4}{RGB}{230,220,255}
\definecolor{GroupColor5}{RGB}{210,200,205}
\definecolor{darkgreen}{HTML}{006400}

\definecolor{AcademicBlue}{HTML}{0055A4} 
\definecolor{ContrastingOrange}{HTML}{E69F00} 
\definecolor{DarkerBlue}{HTML}{002D5A}
 \definecolor{DarkGray}{HTML}{2F2F2F}
\usepackage{pbalance}

\usepackage{newfloat}
\usepackage{listings}
\usepackage{booktabs}
\usepackage{tcolorbox}
\definecolor{CVPRBoxBackground}{rgb}{0.96, 0.94, 0.90} 
\definecolor{CVPRBoxBorder}{rgb}{0.82, 0.75, 0.68}     
\definecolor{CVPRTitle}{rgb}{0.10, 0.10, 0.10}

\newcommand{\modelname}{AVATAR}
\newcommand{\modelwithlogo}{\textbf{\modelname}\! \raisebox{-0.2ex}{\includegraphics[height=0.9em]{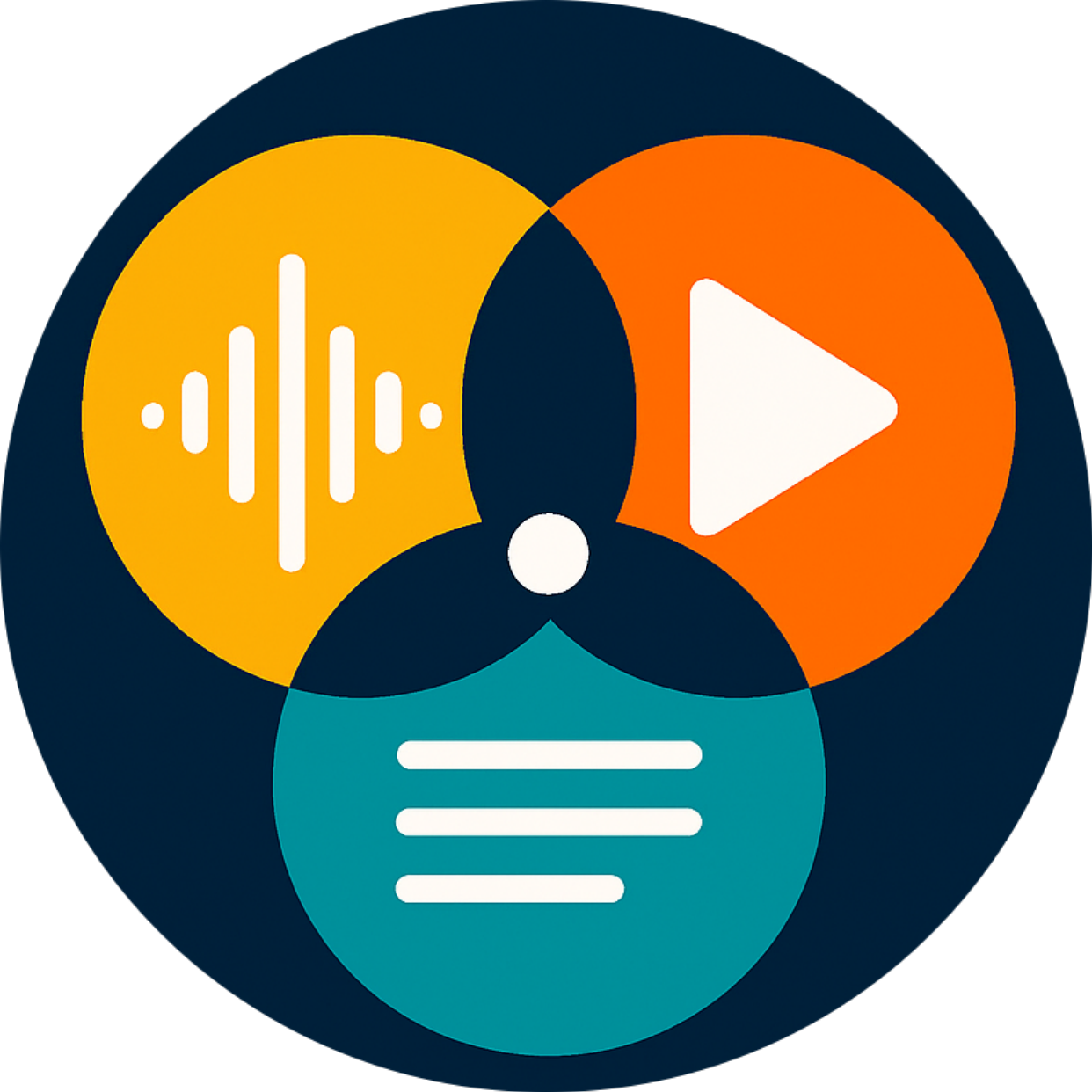}}}

\makeatletter
\renewcommand\paragraph{\@startsection{paragraph}{4}{\z@}
  {0.8ex \@plus0.1pt \@minus.5ex}
  {-0.2em}
  {\normalfont\normalsize\bfseries}}
\makeatother
\makeatletter

\floatstyle{ruled}
\newfloat{listing}{tb}{lst}{}
\floatname{listing}{Listing}

\lstdefinestyle{cleanpythonstyle}{
    language=Python,
    backgroundcolor=\color{gray!5},
    commentstyle=\color{green!60!black}\itshape,
    keywordstyle=\color{blue!80!black}\bfseries,
    numberstyle=\tiny\color{gray!70},
    stringstyle=\color{red!70!black},
    identifierstyle=\color{black},
    basicstyle=\ttfamily\small,
    breakatwhitespace=false,         
    breaklines=true,                 
    postbreak=\mbox{\textcolor{red}{$\hookrightarrow$}\space},
    captionpos=b,                    
    keepspaces=true,                 
    numbers=left,                    
    numbersep=8pt,
    stepnumber=1,
    firstnumber=1,
    showspaces=false,                
    showstringspaces=false,
    showtabs=false,                  
    tabsize=4,
    frame=tb,
    framerule=0.5pt,
    framesep=5pt,
    xleftmargin=15pt,
    xrightmargin=5pt,
    aboveskip=10pt,
    belowskip=10pt,
    escapeinside={(*@}{@*)},
    mathescape=true,
}

\def\paperID{xxxx} 
\def\confName{CVPR}
\def\confYear{2026}

\title{ \hspace{-0.8em}~\raisebox{-1.3ex}{\includegraphics[height=2em]{images/avatar_logo.pdf}}\hspace{0.3em}\modelname{}: Reinforcement Learning to See, Hear, and Reason Over Video}

\author{
    Yogesh Kulkarni \quad Pooyan Fazli \\
    Arizona State University \\
    {\tt\small \{ykulka10, pooyan\}@asu.edu}\\
    {\vspace{0.5em}\textcolor{cvprblue}{\small \url{https://people-robots.github.io/AVATAR/}}}
}

\begin{document}
\maketitle
\begin{abstract}
Multimodal reasoning over long-horizon video is challenging due to the need for precise spatiotemporal fusion and alignment across modalities. While recent methods such as Group Relative Policy Optimization (GRPO) have shown promise in this domain, they suffer from three key limitations: (1) data inefficiency from their on-policy design, (2) a vanishing advantage problem, where identical or near-identical rewards within a group eliminate the learning signal by producing zero-valued advantages, and (3) uniform credit assignment that fails to emphasize critical reasoning steps.
We introduce \textbf{AVATAR} (\textbf{A}udio-\textbf{V}ideo \textbf{A}gen\textbf{t} for \textbf{A}lignment and \textbf{R}easoning), a framework that addresses these limitations through two core components: (1) an off-policy training architecture that improves sample efficiency and resolves vanishing advantages by reusing past experiences with greater reward diversity, and (2) Temporal Advantage Shaping (TAS), a credit assignment strategy that emphasizes early (planning) and late (synthesis) reasoning phases. 
\textbf{AVATAR} achieves strong performance across various benchmarks, outperforming the Qwen2.5-Omni baseline by $\mathbf{+5.4}$ on MMVU, $\mathbf{+4.9}$ on OmniBench, and $\mathbf{+4.5}$ on Video-Holmes. Furthermore, it surpasses standard GRPO by $\mathbf{+3.7}$ on OmniBench and $\mathbf{+1.9}$ on Video-Holmes, while demonstrating \textbf{$5$$\times$ sample efficiency}, requiring $80\%$ fewer generated completions to reach target performance.

\end{abstract}    
\section{Introduction}

Multimodal large language models (MLLMs) must align video, audio, and language modalities to support long-horizon reasoning~\cite{avreasoner}. This requires balancing dense temporal coverage for narrative understanding with high spatial resolution for visual grounding \cite{omnir1, scalingrl, vidhalluc}. Reinforcement learning (RL), particularly Group Relative Policy Optimization (GRPO)~\cite{deepseek, sftorrl}, has emerged as a promising approach for enhancing such reasoning. However, while GRPO excels in verifiable domains such as mathematics~\cite{vlrethinker}, where ground-truth rewards are dense and deterministic, it exhibits significant limitations in open-ended video domains.

\begin{figure}[!t]
\centering
\includegraphics[width=1.0\columnwidth]{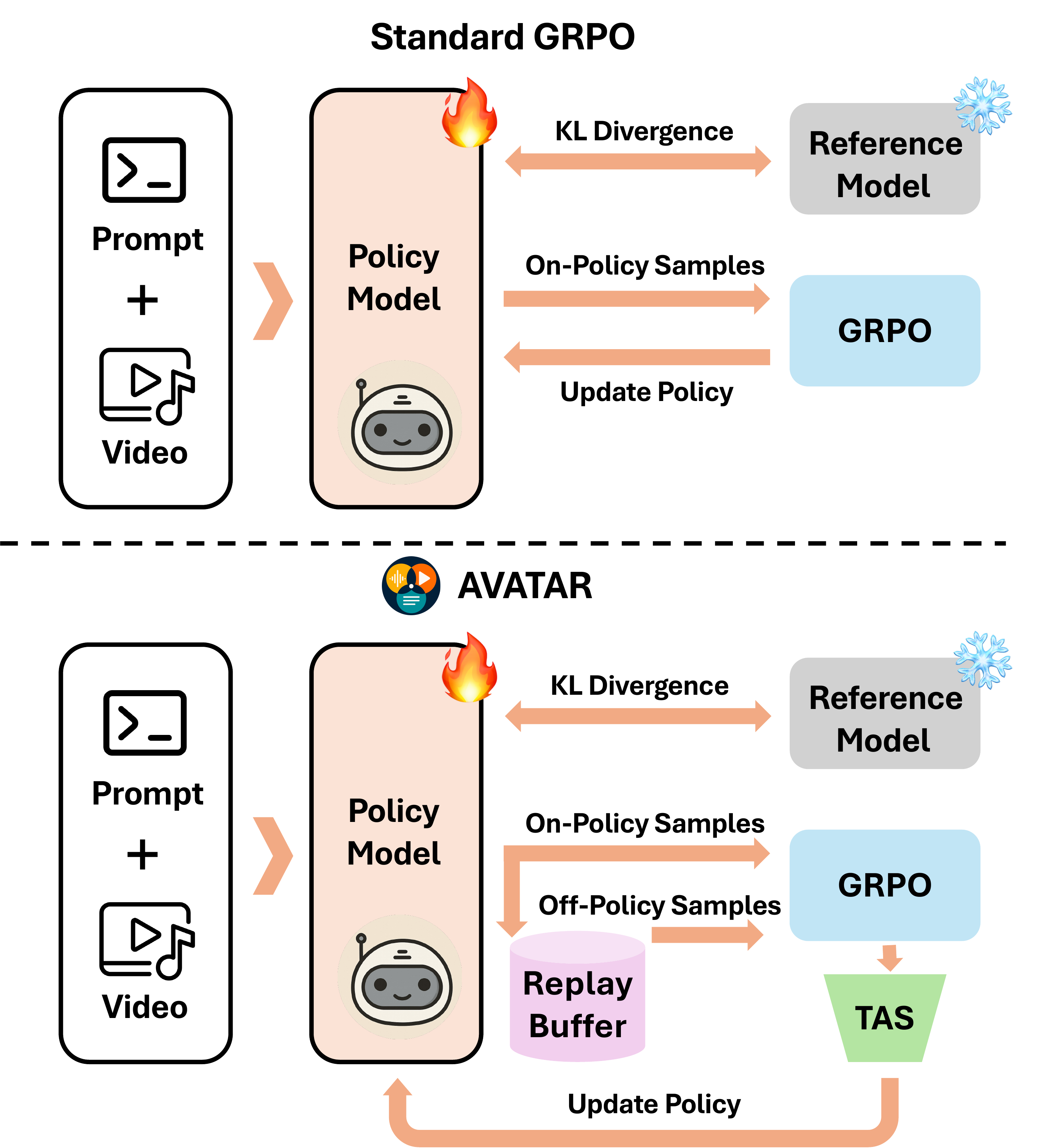}
\caption{\textbf{Standard GRPO (top) vs.\ \modelname{} (bottom).} \modelname{} enhances GRPO with two key components: (1) an \textbf{off-policy architecture} using a stratified replay buffer to improve data efficiency, and (2) \textbf{Temporal Advantage Shaping (TAS)}, a novel credit assignment strategy that focuses learning on critical reasoning steps.}
\label{fig:teaser}
\end{figure}

\textbf{First}, GRPO is an on-policy method that does not reuse past samples, making it inherently data-inefficient. This is particularly problematic in the video domain, where training relies on expensive, large-scale annotated data~\cite{Liu_2025_ICCV, videorft, videorts}. While prior methods such as RePO~\cite{repo, vlrethinker} and standard training frameworks like veRL~\cite{verl} attempt to reuse past data, they often rely on naive replay mechanisms that treat all past experiences as equally informative, failing to prioritize difficult samples and failure cases that provide the strongest learning signals.

\textbf{Second}, GRPO suffers from the vanishing advantage problem: when reward variance within a group collapses, such as when all responses are uniformly correct or incorrect, the resulting zero-valued advantages eliminate the learning signal~\cite{vlrethinker, understandingr1zero, sgrpo, feng2025don}. 
Prior work partially addresses this issue. DAPO~\cite{yu2025dapo} modifies sampling to reduce uniform groups, but remains susceptible to zero gradients on consistently difficult queries where all samples receive identical rewards. Prioritized Experience Replay (PER)~\cite{schaul2015prioritized} prioritizes individual samples by error magnitude but does not ensure sufficient reward variation within training groups, leaving the vanishing advantage problem unresolved.

\textbf{Third}, GRPO suffers from uniform credit assignment~\cite{rlonlyinname, nthr}, assigning the same reward to all tokens in a reasoning chain regardless of their contribution to the final response~\cite{twgrpo, sophia, videocap}. This overlooks key inductive biases in Transformer architectures, including the \textit{attention sink} effect~\cite{xiao2023efficient}, where initial tokens consistently receive attention and act as stable reference points for planning, and the critical role of final tokens in synthesizing the final answer. As a result, GRPO fails to adequately reinforce these critical planning and synthesis phases.

To address these challenges, we introduce \textbf{AVATAR} (\textbf{\underline{A}}udio-\textbf{\underline{V}}ideo \textbf{\underline{A}}gen\textbf{t} for \textbf{\underline{A}}lignment and \textbf{\underline{R}}easoning), a unified framework built on GRPO for multimodal reasoning, with a focus on video tasks. 
To mitigate GRPO’s data inefficiency, \modelname{} adopts an \textbf{off-policy training architecture} that reuses past experiences through a stratified replay buffer. While prior replay-based approaches reuse past data~\cite{schaul2015prioritized, vlrethinker}, our method samples training groups from a replay buffer that retains both successful and failed trajectories based on historical performance. This induces reward variation within each group, ensuring non-zero advantages and enabling gradient updates even for difficult queries where reward variance would otherwise collapse.

We also address the uniform credit assignment problem with a new method called \textbf{Temporal Advantage Shaping (TAS)}. Unlike prior work that treats all tokens equally~\cite{rlonlyinname} or depends on statistical variance~\cite{twgrpo}, TAS guides learning based on a token's position in the reasoning chain, placing greater weight on the beginning (planning) and end (synthesis) stages.
This is especially useful for audio-video tasks, where accurate early grounding (e.g., locating the speaker) and final synthesis (e.g., combining speech and visual cues to identify the speaker) are critical. In the experiments, we apply \modelname{} within a three-stage RL training pipeline to evaluate its effectiveness: starting with general visual reasoning, followed by audio-visual alignment, and concluding with fine-grained audio-based object localization. In summary, our primary contributions are: 
\begin{enumerate}
    \item We present \textbf{\modelname{}}, an off-policy RL framework designed to improve data efficiency in audio-video reasoning by leveraging past experiences through a difficulty-aware replay buffer that prioritizes challenging samples.

     \item We introduce \textbf{Temporal Advantage Shaping (TAS)}, a credit assignment strategy that applies position-dependent weighting to emphasize the planning and synthesis phases of reasoning during training.
    \item Experiments show that \textbf{\modelname{}} achieves significant gains over the Qwen2.5 Omni baseline, with improvements of $+5.4$ on MMVU, $+4.9$ on OmniBench, and $+4.5$ on Video-Holmes. Furthermore, \modelname{} consistently outperforms standard GRPO, showing gains of $+3.7$ on OmniBench and $+1.6$ on MMVU, while requiring $80\%$ fewer generated completions to reach target performance.
\end{enumerate}

\section{Related Work}
\paragraph{Multimodal Understanding.} Recent MLLMs have made remarkable progress towards audio-video understanding~\cite{qwen25vl, internvl3, salmonn, vita, empowering, bimba}. Key efforts focus on architectural enhancements such as dedicated audio branches and video connectors~\cite{videollama2}, using clue aggregators to ground responses in question-specific details~\cite{cat}, or with token interleaving to enhance temporal understanding~\cite{empowering}. However, these methods excel at direct video question answering but cannot rationalize their thoughts through reasoning traces. This limitation is problematic for complex audio-visual tasks requiring multi-step reasoning and fine-grained cross-modal alignment, such as localizing sound sources. In contrast, \modelname{} enhances structured reasoning capabilities in audio-visual models using a sample-efficient RL approach.

\paragraph{Multimodal Understanding with RL.} 

Early multimodal efforts use preference optimization like DPO~\cite{dpo, videopasta, videosavi, rrpo}, but RL methods such as GRPO~\cite{deepseek, sftorrl, kulkarni2025egovita} have shown greater promise for reasoning. Recent variants adapt GRPO to specific tasks: Video-R1 uses temporal contrastive rewards~\cite{videor1}, HumanOmniV2 employs LLM-judged rewards for global audio-visual context~\cite{humanomni}, Omni-R1 targets audio-visual segmentation~\cite{omnir1}, and GRIT generates spatial reasoning chains~\cite{grit}. However, these approaches suffer from uniform credit assignment and sample inefficiency. \modelname{} addresses these limitations through its off-policy architecture and TAS credit assignment. For audio-visual tasks, its temporal weighting emphasizes early grounding (localizing sounds) and late synthesis (combining auditory and visual evidence). Our stratified replay buffer ensures reward diversity within training groups, maintaining a non-zero learning signal even for consistently difficult queries. In contrast, prior methods reuse past experiences but do not explicitly ensure reward diversity within groups~\cite{repo, vlrethinker}.

\section{Preliminaries}
\subsection{Group Relative Policy Optimization}

GRPO is an RL framework that can be used to fine-tune MLLMs by comparing multiple candidate responses generated by a behavior policy $\pi_{\theta_{\text{old}}}$ for a given prompt $q$~\cite{deepseek}. Each candidate output $o_i$ typically includes both intermediate reasoning steps and a final answer. For each group of $K$ candidate responses $\{o_i\}_{i=1}^{K}$, scalar rewards $R(o_i)$ from predefined reward functions are assigned to each response. These rewards are then normalized within the group to obtain relative advantages $A_i$.

\begin{equation}
\label{eq:grpo_advantage}
A_i = \frac{R(o_i) - \mu_R}{\sigma_R + \epsilon_{\mathrm{adv}}},
\end{equation}

\noindent where $\mu_R$ and $\sigma_R$ are the mean and standard deviation of the group rewards $\{R(o_i)\}_{i=1}^K$ and $\epsilon_{adv}$ is a small constant for numerical stability. These group-relative advantages guide the update of the target policy $\pi_\theta$, encouraging it to assign higher probability to relatively better responses. GRPO maximizes the following clipped surrogate objective with KL regularization~\cite{repo}:
\begin{equation}
\begin{split}
    \mathcal{J}_{\mathrm{GRPO}}(\theta)
    &= \mathbb{E}_{\substack{q \sim \mathcal{D} \\ \{o_i\}_{i=1}^K \sim \pi_{\theta_{\mathrm{old}}}}}
    \Bigg[
        \frac{1}{K}\sum_{i=1}^{K}
        \min\!\big(
            r_i A_i,\;
            \bar{r}_i A_i
        \big) \\
    &\qquad
        - \beta\, D_{\mathrm{KL}}\!\big(
            \pi_{\theta}(\cdot|q)
            \,\|\, 
            \pi_{\text{ref}}(\cdot|q)
        \big)
    \Bigg],
\end{split}
\label{eq:grpo_loss}
\end{equation}
\noindent where $\bar{r}_i = \mathrm{clip}(r_i, 1-\epsilon, 1+\epsilon)$, $q$ is a prompt sampled from the dataset $\mathcal{D}$, $r_i$ is the importance sampling ratio $\frac{\pi_{\theta}(o_i | q)}{\pi_{\theta_{\text{old}}}(o_i | q)}$, $\epsilon$ controls the clipping range, and $\beta$ weights the KL divergence term to ensure stability by regularizing the updated policy $\pi_{\theta}$ against a reference policy $\pi_{\mathrm{ref}}$. Here, $\pi_{\mathrm{ref}}$ refers to the pretrained model before RL fine-tuning, used to prevent large deviations from the original policy.

\subsection{GRPO Limitations}
\paragraph{Data Inefficiency Problem.}

GRPO is an on-policy method that discards experiences after a single update, making it data inefficient. This is particularly problematic for audio-visual reasoning, which relies on scarce and costly annotated data. When the pre-trained policy struggles with a complex prompt, it cannot learn from these failures, as the corresponding experiences are discarded rather than reused.

\paragraph{Vanishing Advantage Problem.}  
A fundamental limitation of GRPO arises from its advantage estimation mechanism. When the rewards \( R(o_i) \) for all outputs in a group are identical or nearly identical (e.g., all zero for a particularly difficult prompt or all one for an easy prompt), the group mean baseline \(\mu_R\) equals each reward. As a result, the advantages become zero for all samples \(i\): \(A_i = R(o_i) - \mu_R = 0\). This eliminates the learning signal and prevents the policy from updating. Consequently, GRPO fails to learn from such training samples, a phenomenon referred to as the \textit{vanishing advantage problem}~\cite{vlrethinker, repo}.

\paragraph{Credit Assignment Problem.}
Another key limitation of GRPO is its overly simplistic credit assignment. In Equation~\ref{eq:grpo_loss}, a single scalar advantage $A_i$ is computed for the entire output $o_i$ and applied uniformly to all tokens during policy updates. This ignores the varying importance of different reasoning phases: tokens critical to initial planning and final synthesis receive the same learning signal as less informative intermediate steps. For long-horizon video reasoning, this uniform treatment is particularly detrimental, as it fails to differentiate drift-prone intermediate tokens from the structurally critical boundaries of the reasoning chain, effectively diluting the gradient signal across the entire sequence.

\section{
  \modelname{}
  \texorpdfstring{
    \hspace{-0.2em}~\raisebox{-1.3ex}{\includegraphics[height=2em]{images/avatar_logo.pdf}}\hspace{0.3em}
  }{AVATAR}
}

\begin{algorithm}[!t]
\caption{\modelname{}}
\label{alg:avatar}
\begin{algorithmic}[1]
\STATE \textbf{Input:} Initial policy $\pi_{\theta_{init}}$, Rewards $R$, Hyperparams $\alpha, \lambda_{\text{TAS}}, \beta, \epsilon, K_{on}, K_{off}$.
\STATE \textbf{Initialize:} $\pi_{\theta} \leftarrow \pi_{\theta_{init}}$, $\pi_{\text{ref}} \leftarrow \pi_{\theta_{init}}$, empty Buffer $\mathcal{B}$, VCRS $\overline{R}(q) \leftarrow \{\}$.
\FOR{each training step}
    \STATE Sample prompts $Q_{batch}$; $\pi_{\theta_{old}} \leftarrow \pi_{\theta}$
    \STATE Generate $K_{on}$ on-policy $E_{on}\!=\!\{(q, o, R(o))\}$ using $\pi_{\theta_{old}}$; Compute $A_{on}$ (Eq.~\ref{eq:grpo_advantage}).
    \STATE Retrieve VCRS $\overline{R}(q)$; Sample $K_{off}$ off-policy $E_{off}\!=\!\{(q, o, R(o), \pi_{\theta_{off}})\}$ from $\mathcal{B}$; Compute $A_{off}$ using $\overline{R}(q)$ (Equation~\ref{eq:vcrs}).
    \STATE Compute shaped advantages $A^{\text{TAS}}_{i,t} \leftarrow (A_{on} \cup A_{off}) \cdot w_{i,t}$ using TAS weights $w_{i,t}$ (Eq.~\ref{eq:tas_weights}).
    \STATE Compute $\mathcal{J}_{\text{total}}(\theta)$ using $A^{\text{TAS}}_{on}, A^{\text{TAS}}_{off}$ and importance ratios $r_i^{on}, r_i^{off}$ (Eq.~\ref{eq:combined_loss}).
    \STATE Update policy: $\theta \leftarrow \theta - \eta \nabla_{\theta} \mathcal{J}_{\text{total}}(\theta)$.
    \STATE Update $\mathcal{B}$ with $E_{on}$ (with stratification/hinting ); Update $\overline{R}(q)$ using $E_{on}$.
\ENDFOR
\end{algorithmic}
\end{algorithm}

\noindent We introduce \modelname{}, an off-policy RL framework for multimodal alignment and reasoning, designed to address key limitations of GRPO. To tackle the vanishing advantage problem and resulting data inefficiency, \modelname{} employs an \textit{off-policy} training architecture using a \textit{stratified replay buffer}. To overcome uniform credit assignment, it introduces \textit{Temporal Advantage Shaping (TAS)}, which modulates advantages across the reasoning sequence by upweighting tokens at the beginning and end of the sequence, ensuring that critical planning and final decision steps receive stronger learning signals. Algorithm~\ref{alg:avatar} outlines the procedure.

\subsection{Off-Policy Architecture}
To address both the data inefficiency of on-policy learning and the vanishing advantage problem, \modelname{} employs an off-policy architecture.

\paragraph{Replay Buffer.}

We employ a stratified replay buffer $\mathcal{B}$ to improve sample efficiency via a progressive curriculum. The buffer of size $10k$ is divided into three fixed-capacity tiers based on difficulty: Easy ($25\%$), Medium ($35\%$), and Hard ($40\%$). At each training step, new experiences $(q, o, \pi_{\theta_{\text{off}}})$, required for off-policy importance sampling, are stored. Tier assignment is determined not by individual rewards, but by the prompt's moving average reward $\bar{R}(q)$, which serves as a dynamic difficulty metric. The tier thresholds are dynamic quantiles based on the $\bar{R}(q)$ score distribution, assigning the bottom $40\%$ of experiences to `Hard', the next $35\%$ to `Medium', and the top $25\%$ to `Easy'. As the agent's performance on a prompt $q$ improves, its $\bar{R}(q)$ rises, and new experiences for that prompt are promoted to easier tiers. This design ensures difficult samples (low $\bar{R}(q)$) are retained in the high-capacity `Hard' tier, forcing the model to learn from its failure modes through repeated exposure. Moreover, this prevents rare successful trajectories on difficult prompts from being overwritten by frequent failures, preserving the reward variance required to prevent advantage collapse.

\paragraph{Hinting Mechanism.}

\modelname{} leverages the KL divergence $D_{KL}(\pi_{\theta} ,|, \pi_{\beta})$ between the target policy $\pi_{\theta}$ and behavior policy $\pi_{\beta}$ to monitor policy stability. When a prompt remains hard (low $\bar{R}(q)$) and the policy stops exploring (low KL), a pre-computed \textit{hint} is triggered. Hints (e.g., ``first locate the object making the sound, then count'') are generated by providing the full problem context (video, audio caption, query, and ground truth) to Qwen2.5-VL-72B~\cite{qwen25vl} and are used as additional guidance to help the agent escape local optima. This ensures training remains in a challenging but solvable regime while promoting targeted exploration~\cite{hintgrpo}.

\paragraph{Hybrid Training with Selective Replay.}
At each training step, a batch is formed from both newly generated on-policy samples and off-policy samples from the replay buffer. To correct for policy drift in off-policy data, we apply importance sampling. The full objective for \modelname{} combines the standard on-policy GRPO loss (Equation~\ref{eq:grpo_loss}) with a corrected off-policy term:

\begin{equation}
    \mathcal{J}_{\text{AVATAR}}(\theta) = \mathcal{J}_{\text{on-policy}}(\theta) + \alpha \cdot \mathcal{J}_{\text{off-policy}}(\theta),
    \label{eq:combined_loss}
\end{equation}
where $\alpha$ controls the off-policy contribution. The off-policy objective $\mathcal{J}{\text{off-policy}}$ follows the same form as Equation~\ref{eq:grpo_loss}, but uses an importance sampling ratio based on the behavior policy $\pi{\theta_{\text{off}}}$ that generated the replayed samples:

\begin{equation}
    r_i^{\text{off}}(\theta) = \frac{\pi_{\theta}(o_i|q)}{\pi_{\theta_{off}}(o_i|q)}.
\end{equation}

\begin{figure}[!t]
\centering
\includegraphics[width=\columnwidth]{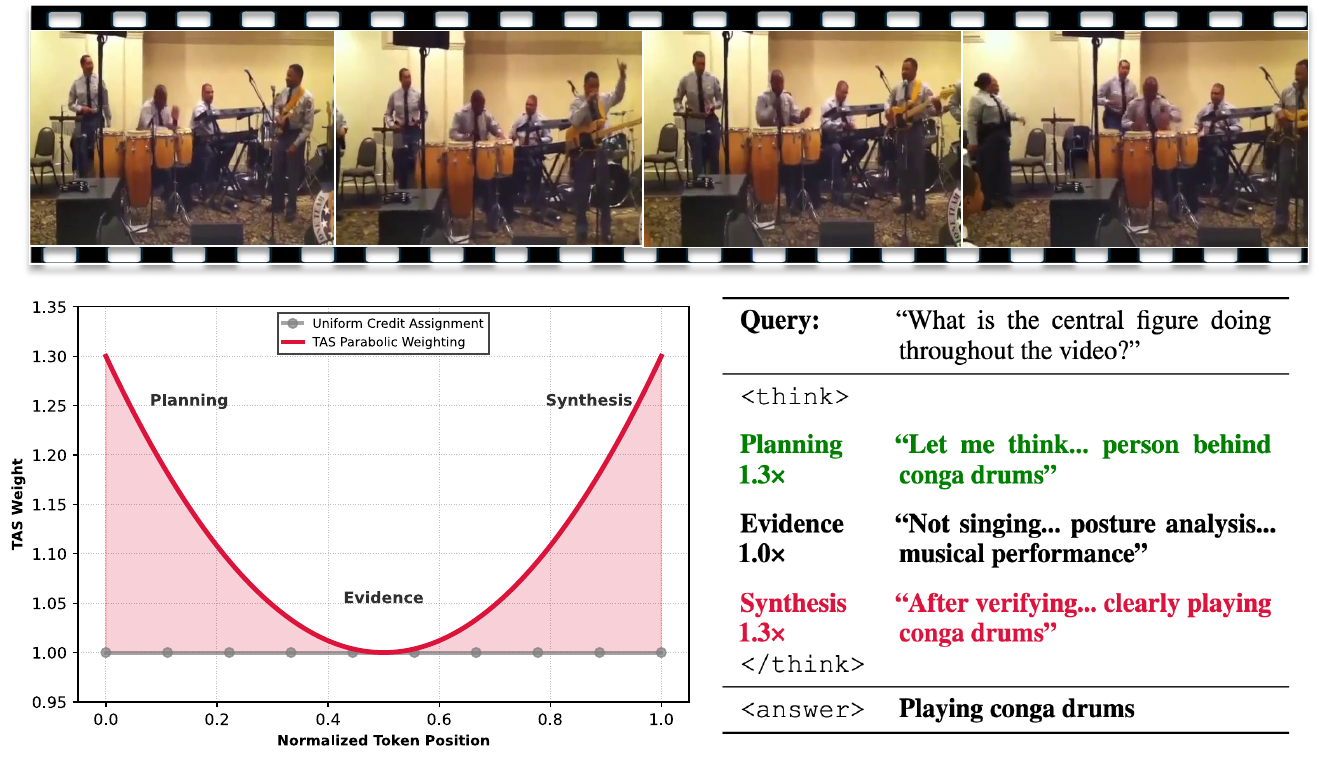}
\caption{To address GRPO’s uniform credit assignment (gray line), TAS applies a parabolic weighting function to amplify advantages during crucial \textbf{planning} and \textbf{synthesis} stages.}
\label{fig:tas}
\end{figure}
\begin{figure*}[t!]
\centering
\includegraphics[width=\textwidth]{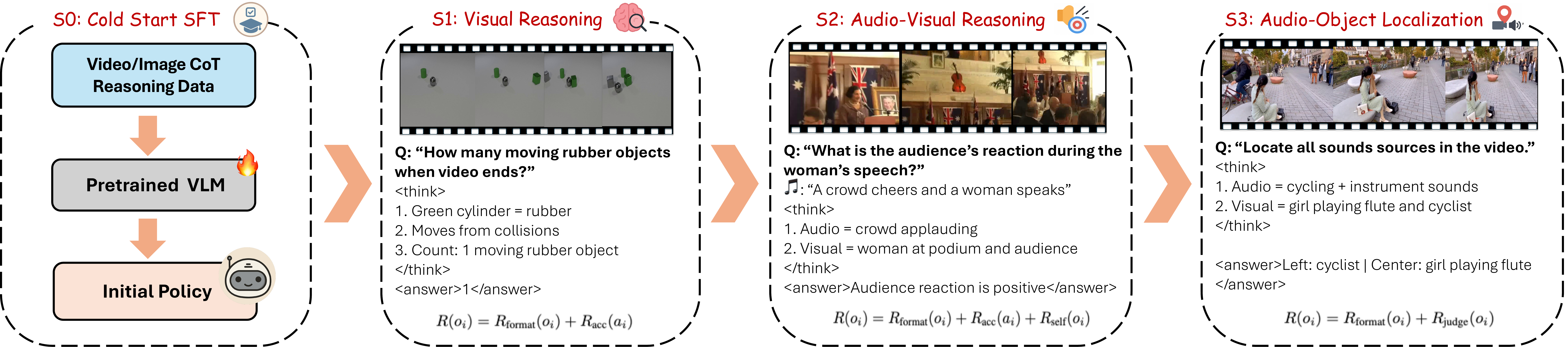}
\caption{
\textbf{Three-stage RL training pipeline to evaluate AVATAR.} The framework advances from \textbf{Cold start SFT} (Stage 0) to \textbf{Visual Reasoning} (Stage 1) to \textbf{Audio-Visual Reasoning} (Stage 2) to \textbf{Audio-Object Localization} (Stage 3). 
}
\label{fig:pipeline}
\end{figure*}

\subsection{Temporal Advantage Shaping (TAS)}

Standard GRPO employs uniform credit assignment, applying the same sequence-level advantage $A_i$ to every token (Equation \ref{eq:grpo_loss}). However, this uniformity is misaligned with the attention patterns of Transformer architectures, including the \textit{attention sink} effect~\cite{xiao2023efficient}, where initial tokens consistently receive attention and act as stable reference points for planning, and the critical role of final tokens in synthesizing the final answer. Motivated by these findings, we introduce Temporal Advantage Shaping (TAS), a simple, critic-free modification to GRPO. As shown in Figure~\ref{fig:tas}, TAS uses a U-shaped parabolic function $w_t$ to yield a token-specific shaped advantage, $A_{i,t}^{TAS} = w_{i,t} \cdot A_i$. This focuses learning on key reasoning phases, reinforcing critical steps without overemphasizing intermediate tokens.

\paragraph{Parabolic Weighting Function.} For a reasoning sequence of length \(L\), we normalize each token’s position \(t \in \{0, 1, \ldots, L-1\}\) to the range \([0,1]\) by defining:
\begin{equation}
\tilde{t} = \frac{t}{L - 1}.
\label{eq:normalization}
\end{equation}
We then compute a token-specific weight using a parabolic function,
\begin{equation}
w_t = 1.0 + \lambda_{\text{TAS}} \cdot (2\tilde{t} - 1)^2,
\label{eq:tas_weights}
\end{equation}
\noindent where \(\lambda_{\text{TAS}}\) controls the amplitude of the weight variation. This function induces a U-shaped weighting over the sequence, assigning minimal weight ($1.0$) to middle tokens (\(\tilde{t} = 0.5\)) and higher weights ($1.0 + \lambda_{\text{TAS}}$) to tokens near the beginning (\(\tilde{t} = 0\)) and end (\(\tilde{t} = 1\)), corresponding to the planning and synthesis phases.

These weights define a token-specific shaped advantage: \(A^{\text{TAS}}_{i,t} = w_{i,t} \cdot A_i\), where $i \in \{1, \dots, K\}$ indexes each of the $K$ candidate responses in the group. \modelname{}’s loss is formed by replacing the standard GRPO advantage $A_i$ in Equation~\ref{eq:grpo_loss} with this shaped advantage. The normalization in Equation~\ref{eq:normalization} ensures that TAS is robust to varying sequence lengths, consistently emphasizing the planning and synthesis phases regardless of generation length.

\subsection{Reward Functions}
\modelname{} is guided by a set of reward functions as detailed below:

\paragraph{(1) Format Reward ($R_{\mathrm{format}}$).}
A binary reward that verifies adherence to the predefined reasoning format (i.e., \texttt{<think>...</think><answer>...</answer>}).
\begin{equation}
    R_{\mathrm{format}}(o_i) = 
    \begin{cases} 
      1, & \text{if format is correct} \\
      -1, & \text{otherwise}
    \end{cases}
\end{equation}

\paragraph{(2) Final Answer Accuracy ($R_{\mathrm{acc}}$).}
A black-box evaluation of the final answer $a_i$ extracted from the \texttt{<answer>} tag, compared to the ground truth $a_{\text{GT}}$. For non-zero numerical tasks, we use the relative Mean Absolute Error (rMAE) for a dense reward signal~\cite{avreasoner}.
\begin{equation}
    R_{\mathrm{acc}}(a_i) = 1.0 - \min\left(1.0, \frac{|a_i - a_{\text{GT}}|}{a_{\text{GT}}}\right)
\end{equation}

\paragraph{(3) Self-Rewarding ($R_{\mathrm{self}}$).}
We use self-rewarding to help the model learn from its own consensus. Given a group of \(K\) generated answers \(\{o_1, o_2, \dots, o_K\}\) for a prompt, we choose the most common answer as a ``pseudo-correct'' \(o^*_{\text{maj}}\) one using majority vote. Ties are broken by choosing the response with the highest model confidence (average token likelihood).

\begin{equation}
    o^*_{\mathrm{maj}} = \mathrm{argmax}_{o \in \{o_1, \dots, o_K\}} \sum_{i=1}^{K} \mathbb{I}(o_i = o),
\end{equation}
where $\mathbb{I}$ is the indicator function. The self-reward for each answer $o_i$ is then a binary score based on its agreement with this consensus:
\begin{equation}
    R_{\mathrm{self}}(o_i) = 
    \begin{cases} 
        1, & \text{if } o_i = o^*_{\mathrm{maj}} \\
        0, & \text{otherwise}
    \end{cases}
\end{equation}

\paragraph{(4) Stepwise Reasoning Judge ($R_{\mathrm{judge}}$).}
A white-box reward from a frozen VLM judge (InternVL3-2B~\cite{internvl3}), which provides a detailed score for reasoning quality within the \texttt{<think>} block. It evaluates logical consistency and correct use of grounding clues from the prompt. The score  $s_{\text{judge}} \in [0, 1]$ reflects the reasoning process quality independently of the final answer.

\begin{table}[!t]
\centering
\caption{Dataset and Reward Configuration for AVATAR Training.}
\label{tab:training_curriculum}
\resizebox{\columnwidth}{!}{
\begin{tabular}{@{}clll@{}}
\toprule
\textbf{Stage} & \textbf{Focus} & \textbf{Datasets} & \textbf{Reward Components} \\
\midrule
\textbf{S0: SFT} & Cold Start & Video-R1-CoT~\cite{videor1}, TPO~\cite{tpo} & N/A (Supervised) \\ \addlinespace
\textbf{S1: RL} & Visual Reasoning & Video-R1~\cite{videor1} & $0.5 \times R_{\text{format}} + 0.5 \times R_{\text{acc}}$ \\ \addlinespace
\textbf{S2: RL} & Audio-Visual Reasoning & AVQA~\cite{avqa}, AVE~\cite{ave} & $0.2 \times R_{\text{format}} + 0.4 \times R_{\text{acc}} + 0.4 \times R_{\text{self}}$ \\ \addlinespace
\textbf{S3: RL} & Audio-Object Localization & AVSBench~\cite{avsbench}& $0.2 \times R_{\text{format}} + 0.4 \times R_{\text{acc}} + 0.4 \times R_{\text{judge}}$ \\
\bottomrule
\end{tabular}
}
\end{table}
\begin{table*}[t!]
\centering
\caption{\textbf{\modelname{} vs. state-of-the-art audio-video understanding models.} Best scores within each RL block are in \textbf{bold}. Performance improvements from applying \modelname{} to a baseline model are shown in \textcolor{darkgreen}{green} with 95\% confidence interval (CI) margins ($\pm$) via  bootstrap. $^{\ddagger}$ Improvement not statistically significant at the 0.05 level. All results were reproduced by us; * denotes potential data contamination.}
\label{tab:av_reasoning_benchmarks}
\begin{adjustbox}{width=\textwidth,center}
\renewcommand{\arraystretch}{1.2}
\fontsize{7pt}{8pt}\selectfont 
\setlength{\tabcolsep}{2mm} 
\begin{tabular}{lcccccc}
\toprule
\makecell{\textbf{Model}} & \makecell{\textbf{OmniBench}} & \makecell{\textbf{DailyOmni}} & \makecell{\textbf{AV-Counting}} & \makecell{\textbf{AV-Odyssey}} & \makecell{\textbf{WorldSense}} & \makecell{\textbf{IntentBench}} \\
\midrule

\rowcolor{gray!15}\multicolumn{7}{c}{\textit{State-of-the-Art Models}}\\
Echolnk~\cite{echoink} & 46.5 & 46.2 & 22.7 & 31.1 & 45.7 & 63.6 \\
Omni-R1~\cite{omnir1} & 46.9 & 46.8 & 22.0 & 31.2 & 44.1 & 63.5 \\
HumanOmni~\cite{humanomni} & 44.9 & 47.6 & 19.6 & 30.3 & 45.4 & * \\
AV-Reasoner~\cite{avreasoner} & 48.3 & 53.8 & 23.0 & 25.6 & 44.6 & 59.5 \\
\midrule

\rowcolor{gray!15}\multicolumn{7}{c}{\textit{Model-Agnostic Reinforcement Learning with \modelname{}}} \\
Ola-7B~\cite{ola} (Baseline) & 45.3 & 52.3 & 17.4 & 25.6 & 44.2 & 59.1 \\
\hspace{3mm} + GRPO & 46.8 $\scriptstyle\textcolor{darkgreen}{(\textbf{+1.5}\pm\textbf{0.4})}$ & 54.1 $\scriptstyle\textcolor{darkgreen}{(\textbf{+1.8}\pm\textbf{0.5})}$ & 18.2 $\scriptstyle\textcolor{darkgreen}{(\textbf{+0.8}\pm\textbf{0.3})}$ & 27.0 $\scriptstyle\textcolor{darkgreen}{(\textbf{+1.4}\pm\textbf{0.5})}$ & 44.7 $\scriptstyle\textcolor{darkgreen}{(\textbf{+0.5}\pm\textbf{0.3})}$ & 60.3 $\scriptstyle\textcolor{darkgreen}{(\textbf{+1.2}\pm\textbf{0.5})}$ \\

\rowcolor{PastaYellow} \hspace{3mm} + \modelwithlogo & \textbf{47.2} $\scriptstyle\textcolor{darkgreen}{(\textbf{+1.9}\pm\textbf{0.5})}$ & \textbf{55.7} $\scriptstyle\textcolor{darkgreen}{(\textbf{+3.4}\pm\textbf{0.7})}$ & \textbf{19.5} $\scriptstyle\textcolor{darkgreen}{(\textbf{+2.1}\pm\textbf{0.6})}$ & \textbf{28.8} $\scriptstyle\textcolor{darkgreen}{(\textbf{+3.2}\pm\textbf{0.6})}$ & \textbf{45.0} $\scriptstyle\textcolor{darkgreen}{(\textbf{+0.8}\pm\textbf{0.4})}$ & \textbf{61.9} $\scriptstyle\textcolor{darkgreen}{(\textbf{+2.8}\pm\textbf{0.6})}$ \\
\midrule

Qwen2.5-Omni~\cite{qwen25omni} (Baseline) & 44.2 & 44.0 & 22.3 & 29.8 & 44.2 & 63.7 \\
\hspace{3mm} + GRPO & 45.4 $\scriptstyle\textcolor{darkgreen}{(\textbf{+1.2}\pm\textbf{0.4})}$ & 44.8 $\scriptstyle\textcolor{darkgreen}{(\textbf{+0.8}\pm\textbf{0.4})}$ & 22.8 $\scriptstyle\textcolor{darkgreen}{(\textbf{+0.5}\pm\textbf{0.3})}$ & 31.3 $\scriptstyle\textcolor{darkgreen}{(\textbf{+1.5}\pm\textbf{0.5})}$ & 45.1 $\scriptstyle\textcolor{darkgreen}{(\textbf{+0.9}\pm\textbf{0.4})}$ & 63.8 $\scriptstyle\textcolor{gray}{(\textbf{+0.1}\pm\textbf{0.1})}^{\ddagger}$ \\

\rowcolor{PastaYellow} \hspace{3mm} + \modelwithlogo & \textbf{49.1} $\scriptstyle\textcolor{darkgreen}{(\textbf{+4.9}\pm\textbf{0.7})}$ & \textbf{47.0} $\scriptstyle\textcolor{darkgreen}{(\textbf{+3.0}\pm\textbf{0.6})}$ & \textbf{23.1} $\scriptstyle\textcolor{darkgreen}{(\textbf{+0.8}\pm\textbf{0.4})}$ & \textbf{32.1} $\scriptstyle\textcolor{darkgreen}{(\textbf{+2.3}\pm\textbf{0.6})}$ & \textbf{46.0} $\scriptstyle\textcolor{darkgreen}{(\textbf{+1.8}\pm\textbf{0.5})}$ & \textbf{63.9} $\scriptstyle\textcolor{darkgreen}{(\textbf{+0.2}\pm\textbf{0.1})}$ \\
\bottomrule
\end{tabular}
\end{adjustbox}
\end{table*}
\begin{table*}[t!]
\centering
\caption{\textbf{\modelname{} on video understanding and reasoning benchmarks.} Best scores within each RL block are in \textbf{bold}. Performance improvements from applying \modelname{} to a baseline model are shown in \textcolor{darkgreen}{green} with 95\% confidence interval (CI) margins ($\pm$) via bootstrap. $^{\ddagger}$ Improvement not statistically significant at the 0.05 level. All results were reproduced by us.}
\label{tab:video_qa_benchmarks}
\begin{adjustbox}{width=\textwidth,center}
\renewcommand{\arraystretch}{1.2}
\fontsize{7pt}{8pt}\selectfont
\setlength{\tabcolsep}{2.5mm}
\begin{tabular}{lcccccc}
\toprule
& \multicolumn{3}{c}{\textbf{General Video Understanding}} & \multicolumn{3}{c}{\textbf{Video Reasoning}} \\
\cmidrule(lr){2-4} \cmidrule(lr){5-7}
\textbf{Model} & \textbf{MVBench} & \textbf{Video-MME} & \textbf{LVBench} & \textbf{Video-Holmes} & \textbf{MMVU} & \textbf{TOMATO} \\
\midrule

\rowcolor{gray!15}\multicolumn{7}{c}{\textit{State-of-the-Art Models}}\\
Echolnk~\cite{echoink} & 66.2 & 60.8 & 37.6 & 42.5 & 65.7 & 29.9 \\
Omni-R1~\cite{omnir1} & 66.0 & 60.7 & 37.6 & 44.2 & 63.6 & 29.2 \\
HumanOmni~\cite{humanomni} & 61.4 & 63.1 & 36.2 & 39.6 & 61.8 & 27.1 \\
AV-Reasoner~\cite{avreasoner} & 47.9 & 56.8 & 33.7 & 39.6 & 57.9 & 24.9 \\
Qwen2.5-VL~\cite{qwen25vl} & 59.4 & 60.1 & 31.2 & 38.0 & 61.9 & 29.3\\
Video-R1~\cite{videor1} & 63.9 & 59.3 & 27.8 & 41.0 & 63.1 & 19.8 \\
VideoRFT~\cite{videorft} & 62.1 & 59.8 & 34.2 & 40.5 & 58.0 & 20.6 \\
TW-GRPO~\cite{twgrpo} & 63.3 & 60.1 & 33.8 & 38.4 & 61.4 & 23.5 \\
\midrule

\rowcolor{gray!15}\multicolumn{7}{c}{\textit{Model-Agnostic Reinforcement Learning with \modelname{}}} \\
Ola-7B~\cite{ola} (Baseline) & 40.1 & 59.1 & 35.5 & 40.1 & 56.6 & 25.3 \\
\hspace{3mm} + GRPO & 42.5 $\scriptstyle\textcolor{darkgreen}{(\textbf{+2.4}\pm\textbf{0.6})}$ & 60.2 $\scriptstyle\textcolor{darkgreen}{(\textbf{+1.1}\pm\textbf{0.4})}$ & 36.0 $\scriptstyle\textcolor{darkgreen}{(\textbf{+0.5}\pm\textbf{0.3})}$ & 41.3 $\scriptstyle\textcolor{darkgreen}{(\textbf{+1.2}\pm\textbf{0.5})}$ & 57.0 $\scriptstyle\textcolor{darkgreen}{(\textbf{+0.4}\pm\textbf{0.3})}$ & 25.9 $\scriptstyle\textcolor{darkgreen}{(\textbf{+0.6}\pm\textbf{0.4})}$ \\

\rowcolor{PastaYellow} \hspace{3mm} + \modelwithlogo & \textbf{45.4} $\scriptstyle\textcolor{darkgreen}{(\textbf{+5.3}\pm\textbf{0.8})}$ & \textbf{61.4} $\scriptstyle\textcolor{darkgreen}{(\textbf{+2.3}\pm\textbf{0.6})}$ & \textbf{36.6} $\scriptstyle\textcolor{darkgreen}{(\textbf{+1.1}\pm\textbf{0.5})}$ & \textbf{42.4} $\scriptstyle\textcolor{darkgreen}{(\textbf{+2.3}\pm\textbf{0.6})}$ & \textbf{57.3} $\scriptstyle\textcolor{darkgreen}{(\textbf{+0.7}\pm\textbf{0.4})}$ & \textbf{26.6} $\scriptstyle\textcolor{darkgreen}{(\textbf{+1.3}\pm\textbf{0.5})}$ \\
\midrule

Qwen2.5-Omni~\cite{qwen25omni} (Baseline) & 66.1 & 58.3 & 37.2 & 40.6 & 60.2 & 29.0 \\
\hspace{3mm} + GRPO & 66.3 $\scriptstyle\textcolor{darkgreen}{(\textbf{+0.2}\pm\textbf{0.1})}$ & 60.5 $\scriptstyle\textcolor{darkgreen}{(\textbf{+2.2}\pm\textbf{0.6})}$ & 37.8 $\scriptstyle\textcolor{darkgreen}{(\textbf{+0.6}\pm\textbf{0.4})}$ & 43.2 $\scriptstyle\textcolor{darkgreen}{(\textbf{+2.6}\pm\textbf{0.6})}$ & 64.0 $\scriptstyle\textcolor{darkgreen}{(\textbf{+3.8}\pm\textbf{0.7})}$ & 29.2 $\scriptstyle\textcolor{gray}{(\textbf{+0.2}\pm\textbf{0.3})}^{\ddagger}$ \\

\rowcolor{PastaYellow} \hspace{3mm} + \modelwithlogo & \textbf{66.4} $\scriptstyle\textcolor{gray}{(\textbf{+0.3}\pm\textbf{0.3})}^{\ddagger}$ & \textbf{62.8} $\scriptstyle\textcolor{darkgreen}{(\textbf{+4.5}\pm\textbf{0.7})}$ & \textbf{38.4} $\scriptstyle\textcolor{darkgreen}{(\textbf{+1.2}\pm\textbf{0.5})}$ & \textbf{45.1} $\scriptstyle\textcolor{darkgreen}{(\textbf{+4.5}\pm\textbf{0.7})}$ & \textbf{65.6} $\scriptstyle\textcolor{darkgreen}{(\textbf{+5.4}\pm\textbf{0.8})}$ & \textbf{30.8} $\scriptstyle\textcolor{darkgreen}{(\textbf{+1.8}\pm\textbf{0.5})}$ \\
\bottomrule
\end{tabular}
\end{adjustbox}
\end{table*}
\begin{table*}[t!]
\centering
\caption{\textbf{Component-wise ablation} demonstrating how each component addresses specific GRPO limitations. The \textbf{stratified replay buffer} resolves data inefficiency and vanishing advantages, \textbf{TAS} improves credit assignment, and \textbf{hinting} helps escape local optima.}
\label{tab:component_ablation}
\begin{adjustbox}{width=\textwidth,center}
\renewcommand{\arraystretch}{1.2}
\fontsize{7pt}{8pt}\selectfont
\setlength{\tabcolsep}{3mm}
\begin{tabular}{lcccccc}
\toprule
& \multicolumn{3}{c}{\textbf{Audio-Visual}} & \multicolumn{3}{c}{\textbf{Video Reasoning}} \\
\cmidrule(lr){2-4} \cmidrule(lr){5-7}
\textbf{Model Configuration} & \textbf{OmniBench} & \textbf{DailyOmni} & \textbf{AV-Odyssey} & \textbf{Video-MMMU} & \textbf{VSI-Bench} & \textbf{Video-TT} \\
\midrule
Qwen2.5 Omni & 44.2 & 44.0 & 29.8 & 46.8 & 25.4 & 41.8 \\
\midrule
+ GRPO & 45.4 $\scriptstyle\textcolor{darkgreen}{(\textbf{+1.2}\pm\textbf{0.4})}$ & 44.8 $\scriptstyle\textcolor{darkgreen}{(\textbf{+0.8}\pm\textbf{0.4})}$ & 31.3 $\scriptstyle\textcolor{darkgreen}{(\textbf{+1.5}\pm\textbf{0.5})}$ & 48.1 $\scriptstyle\textcolor{darkgreen}{(\textbf{+1.3}\pm\textbf{0.5})}$ & 25.9 $\scriptstyle\textcolor{darkgreen}{(\textbf{+0.5}\pm\textbf{0.3})}$ & 43.0 $\scriptstyle\textcolor{darkgreen}{(\textbf{+1.2}\pm\textbf{0.5})}$ \\
+ TAS Only (w/ On-Policy GRPO) & 45.1 $\scriptstyle\textcolor{darkgreen}{(\textbf{+0.9}\pm\textbf{0.4})}$ & 45.4 $\scriptstyle\textcolor{darkgreen}{(\textbf{+1.4}\pm\textbf{0.5})}$ & 31.4 $\scriptstyle\textcolor{darkgreen}{(\textbf{+1.6}\pm\textbf{0.5})}$ & 49.0 $\scriptstyle\textcolor{darkgreen}{(\textbf{+2.2}\pm\textbf{0.6})}$ & 26.5 $\scriptstyle\textcolor{darkgreen}{(\textbf{+1.1}\pm\textbf{0.3})}$ & 43.8 $\scriptstyle\textcolor{darkgreen}{(\textbf{+2.0}\pm\textbf{0.6})}$ \\
+ Replay Buffer (w/ Uniform Credit)& 47.8 $\scriptstyle\textcolor{darkgreen}{(\textbf{+3.6}\pm\textbf{0.6})}$ & 45.9 $\scriptstyle\textcolor{darkgreen}{(\textbf{+1.9}\pm\textbf{0.5})}$ & 31.6 $\scriptstyle\textcolor{darkgreen}{(\textbf{+1.8}\pm\textbf{0.5})}$ & 48.5 $\scriptstyle\textcolor{darkgreen}{(\textbf{+1.7}\pm\textbf{0.5})}$ & 26.1 $\scriptstyle\textcolor{darkgreen}{(\textbf{+0.7}\pm\textbf{0.3})}$ & 43.3 $\scriptstyle\textcolor{darkgreen}{(\textbf{+1.5}\pm\textbf{0.5})}$ \\
\midrule
\rowcolor{PastaYellow} \textbf{\modelname{}} & \textbf{49.1} $\scriptstyle\textcolor{darkgreen}{(\textbf{+4.9}\pm\textbf{0.7})}$ & \textbf{47.0} $\scriptstyle\textcolor{darkgreen}{(\textbf{+3.0}\pm\textbf{0.6})}$ & \textbf{32.1} $\scriptstyle\textcolor{darkgreen}{(\textbf{+2.3}\pm\textbf{0.6})}$ & \textbf{49.4} $\scriptstyle\textcolor{darkgreen}{(\textbf{+2.6}\pm\textbf{0.7})}$ & \textbf{26.8} $\scriptstyle\textcolor{darkgreen}{(\textbf{+1.4}\pm\textbf{0.4})}$ & \textbf{44.2} $\scriptstyle\textcolor{darkgreen}{(\textbf{+2.4}\pm\textbf{0.7})}$ \\
\bottomrule
\end{tabular}
\end{adjustbox}
\end{table*}

\subsection{Video-Context Reference Score (VCRS)}
Standard off-policy updates can be noisy, as the group mean ($\mu_R$) is an unstable baseline for stale buffer samples. We stabilize learning by introducing VCRS, denoted as $\bar{R}(q)$, which is a stable moving average of rewards for a prompt $q$ computed over its last 20 new (on-policy) instances.

This stable score $\bar{R}(q)$ replaces the noisy $\mu_R$ in the off-policy advantage calculation:
\begin{equation}
\label{eq:vcrs}
A_{i,\text{off}} = \frac{R(o_i) - \bar{R}(q)}{\sigma_{R,\text{off}}},
\end{equation}
where $\sigma_{R,\text{off}}$ is the standard deviation of the off-policy batch. This provides a reliable advantage signal, advantage $A_{i,\text{off}}$, which is modulated by TAS.

\section{Experiments}
\label{sec:experiments}

\paragraph{Implementation Details.}
We evaluate \modelname{} on Qwen2.5-Omni-7B~\cite{qwen25omni} and Ola-7B~\cite{ola}, two open-source MLLMs that provide native audio and video support and are the current SOTA in this space. We implement \modelname{} using the MS-Swift~\cite{swift} framework for GRPO training. Our training follows the curriculum detailed in Table~\ref{tab:training_curriculum} and Figure~\ref{fig:pipeline}, beginning with a cold-start SFT phase (S0) followed by three RL stages (S1-S3) of increasing complexity, each targeting different reasoning skills using specific datasets and reward configurations. Stage 2 uses audio cues annotated via Kimi-Audio~\cite{kimi}, and Stage 3 employs stepwise rewards from an InternVL3~\cite{internvl3} judge. We uniformly sample 32 video frames during both training and evaluation. TAS is applied throughout all RL stages. We attach LoRA~\cite{hu2021lora} adapters to the Vision Encoder, Audio Encoder, MLP, and the LLM with parameters $r = 16$ and $\alpha_{\text{LoRA}} = 32$. The key hyperparameters for \modelname{} are \( \lambda_{TAS} = 0.3, \alpha = 0.6, \beta = 0.1 \).

\paragraph{State-of-the-art Models.}
We compare \modelname{} with a broad range of models across audio-visual and video-only modalities. For audio-visual models, we evaluate against Qwen2.5-Omni-7B~\cite{qwen25omni}, EchoInk~\cite{echoink}, Omni-R1~\cite{omnir1}, HumanOmni~\cite{humanomni}, Ola-7B~\cite{ola}, and AV-Reasoner~\cite{avreasoner}. To assess the impact of audio integration, we also compare with SOTA video-only models: Qwen2.5VL~\cite{qwen25vl}, Video-R1~\cite{videor1}, Video-RFT~\cite{videorft}, and TW-GRPO~\cite{twgrpo}.

\paragraph{Benchmarks.}
For audio-visual tasks, we use OmniBench~\cite{omnibench} (image-audio-text reasoning), DailyOmni~\cite{dailyomni} (ambient sound scenarios), AV-Counting~\cite{avreasoner} (cross-modal counting), AV-Odyssey~\cite{avodyssey} (audio-visual alignment), WorldSense~\cite{worlsense} (omnimodal understanding in-the-wild), and IntentBench~\cite{humanomni} (reasoning about human intent and actions). For video understanding and reasoning, we evaluate on MVBench~\cite{mvbench} (temporal perception-to-cognition QA), Video-MME~\cite{videomme} (full-spectrum video comprehension), LVBench~\cite{lvbench} (long-video reasoning), Video-Holmes~\cite{videoholmes} (causal inference in suspense clips), MMVU~\cite{mmvu} (domain-expert knowledge analysis), and TOMATO~\cite{tomato} (temporal reasoning).

\begin{figure}[!t]
    \centering
    \begin{subfigure}[b]{0.48\columnwidth}
        \centering
        \includegraphics[width=\textwidth]{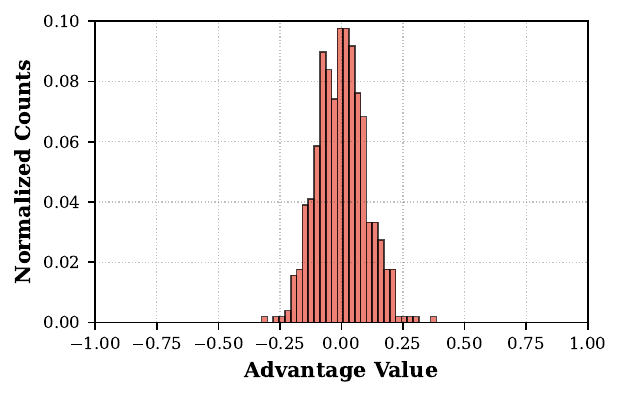}
        \caption{GRPO Advantages Dist.}
    \end{subfigure}
    \hfill
    \begin{subfigure}[b]{0.48\columnwidth}
        \centering
        \includegraphics[width=\textwidth]{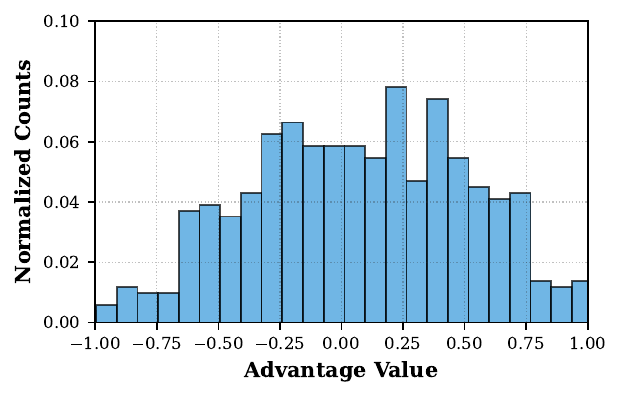}
        \caption{\modelname{} Advantages Dist.}
    \end{subfigure}
\caption{Advantage distribution comparison. (a) GRPO \textbf{exhibits vanishing advantages} with concentration around zero. (b) AVATAR \textbf{maintains diverse advantages} through replay buffer.}
    
    \label{fig:advantages}
\end{figure}

\subsection{Results}

\paragraph{Audio-Visual Reasoning.}
As shown in Table~\ref{tab:av_reasoning_benchmarks}, \modelname{} outperforms both the baseline and standard GRPO across all benchmarks. On Qwen2.5-Omni, \modelname{} achieves $\mathbf{+4.9}$ on OmniBench (vs.\ GRPO's $+1.2$), $\mathbf{+3.0}$ on DailyOmni (vs.\ $+0.8$), and $\mathbf{+2.3}$ on AV-Odyssey (vs.\ $+1.5$), nearly $4\times$ the gain of GRPO on OmniBench. On Ola-7B, \modelname{} similarly doubles GRPO's improvements, notably $\mathbf{+3.4}$ vs.\ $+1.8$ on DailyOmni and $\mathbf{+3.2}$ vs.\ $+1.4$ on AV-Odyssey. These gains confirm that the Stratified Replay Buffer resolves the vanishing advantage problem by reconstructing group-wise reward variance, while TAS prevents cross-modal attention drift by upweighting critical planning and synthesis tokens.

\paragraph{General Video Understanding and Reasoning.}

\modelname{} demonstrates strong performance on reasoning-heavy benchmarks in Table~\ref{tab:video_qa_benchmarks}, showing improved stability compared to standard methods. On Video-Holmes, \modelname{} nearly doubles the gain of GRPO with $\mathbf{+4.5}$ compared to $+2.6$. On TOMATO for temporal ordering, standard GRPO does not yield statistically significant improvements, whereas \modelname{} achieves a significant $\mathbf{+1.8}$ gain. This also highlights the role of TAS: while uniform weighting methods such as TW-GRPO~\cite{twgrpo} apply equal credit across the sequence, TAS applies position-dependent weighting, concentrating optimization on the planning and synthesis phases.

\begin{figure}[t!]
    \centering
    \begin{subfigure}[b]{0.48\columnwidth}
        \centering
        \includegraphics[width=\textwidth]{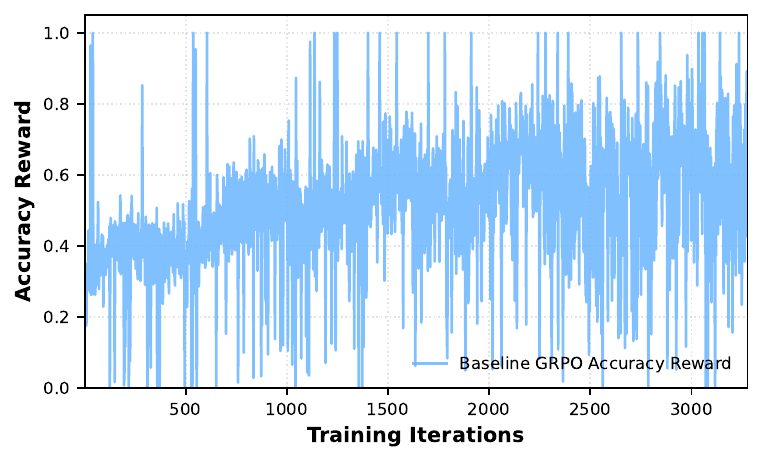}
        \caption{GRPO}
        \label{fig:baseline}
    \end{subfigure}
    \hfill
    \begin{subfigure}[b]{0.48\columnwidth}
        \centering
        \includegraphics[width=\textwidth]{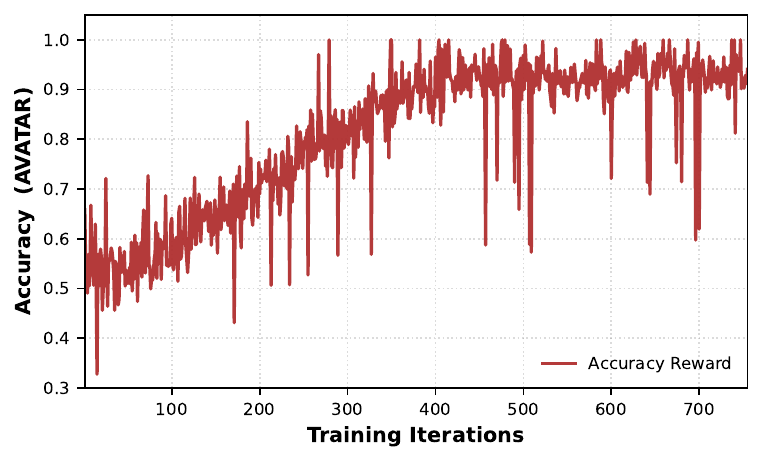}
        \caption{\modelname{}}
        \label{fig:avatar}
    \end{subfigure}
\caption{
Comparison of training dynamics between GRPO and \modelname{}. GRPO (a) shows \textbf{oscillatory and unstable accuracy reward progression}, whereas \modelname{} (b) demonstrates a \textbf{smoother, more consistent learning trajectory}.
}
    \label{fig:training_comparison}
\end{figure}
\begin{table*}[!t]
\centering
\caption{\textbf{Ablation studies on training curriculum and advantage shaping strategies}. \textbf{TAS} significantly outperforms uniform weights and inverse parabolic weights, validating our hypothesis that \textbf{early} and \textbf{late} reasoning phases are most critical.}
\label{tab:curriculum_reward_ablation}
\begin{adjustbox}{width=\textwidth,center}
\renewcommand{\arraystretch}{1.2}
\fontsize{7pt}{8pt}\selectfont
\setlength{\tabcolsep}{2mm}
\begin{tabular}{ll ccc ccc}
\toprule
\multicolumn{2}{c}{\textbf{Ablation Configuration}} & \multicolumn{3}{c}{\textbf{Audio-Visual}} & \multicolumn{3}{c}{\textbf{Video Reasoning}} \\
\midrule
\textbf{Group} & \textbf{Setting} & \textbf{OmniBench} & \textbf{DailyOmni} & \textbf{AV-Odyssey} & \textbf{Video-Holmes} & \textbf{MMVU} & \textbf{TOMATO} \\
\midrule
Qwen2.5 Omni &  & 44.2 & 44.0 & 29.8 & 40.6 & 60.2 & 29.0 \\
\midrule
\multirow{4}{*}{\textit{Training Curriculum}} 
& SFT Only 
& 45.8 $\scriptstyle\textcolor{darkgreen}{(\textbf{+1.6})}$ 
& 45.2 $\scriptstyle\textcolor{darkgreen}{(\textbf{+1.2})}$ 
& 30.1 $\scriptstyle\textcolor{darkgreen}{(\textbf{+0.3})}$ 
& 41.8 $\scriptstyle\textcolor{darkgreen}{(\textbf{+1.2})}$ 
& 62.1 $\scriptstyle\textcolor{darkgreen}{(\textbf{+1.9})}$ 
& 28.9 $\scriptstyle\textcolor{darkgreen}{(\textbf{-0.1})}$ \\
& SFT + Stage 1 RL 
& 47.2 $\scriptstyle\textcolor{darkgreen}{(\textbf{+3.0})}$ 
& 45.8 $\scriptstyle\textcolor{darkgreen}{(\textbf{+1.8})}$ 
& 30.6 $\scriptstyle\textcolor{darkgreen}{(\textbf{+0.8})}$ 
& 43.5 $\scriptstyle\textcolor{darkgreen}{(\textbf{+2.9})}$ 
& 64.2 $\scriptstyle\textcolor{darkgreen}{(\textbf{+4.0})}$ 
& 29.4 $\scriptstyle\textcolor{darkgreen}{(\textbf{+0.4})}$ \\
& SFT + Stages 1+2 RL 
& 48.6 $\scriptstyle\textcolor{darkgreen}{(\textbf{+4.4})}$ 
& 46.7 $\scriptstyle\textcolor{darkgreen}{(\textbf{+2.7})}$ 
& 31.8 $\scriptstyle\textcolor{darkgreen}{(\textbf{+2.0})}$ 
& 44.2 $\scriptstyle\textcolor{darkgreen}{(\textbf{+3.6})}$ 
& 65.1 $\scriptstyle\textcolor{darkgreen}{(\textbf{+4.9})}$ 
& 30.1 $\scriptstyle\textcolor{darkgreen}{(\textbf{+1.1})}$ \\
 \rowcolor{PastaYellow}& \textbf{SFT + Stages 1+2+3 RL (Ours)} 
& \textbf{49.1} $\scriptstyle\textcolor{darkgreen}{(\textbf{+4.9})}$ 
& \textbf{47.0} $\scriptstyle\textcolor{darkgreen}{(\textbf{+3.0})}$ 
& \textbf{32.1} $\scriptstyle\textcolor{darkgreen}{(\textbf{+2.3})}$ 
& \textbf{45.1} $\scriptstyle\textcolor{darkgreen}{(\textbf{+4.5})}$ 
& \textbf{65.6} $\scriptstyle\textcolor{darkgreen}{(\textbf{+5.4})}$ 
& \textbf{30.8} $\scriptstyle\textcolor{darkgreen}{(\textbf{+1.8})}$ \\
\midrule
\multirow{5}{*}{\textit{Advantage Shaping}} 
& Linear Decay Weights 
& 48.3 $\scriptstyle\textcolor{darkgreen}{(\textbf{+4.1})}$ 
& 46.2 $\scriptstyle\textcolor{darkgreen}{(\textbf{+2.2})}$ 
& 31.6 $\scriptstyle\textcolor{darkgreen}{(\textbf{+1.8})}$ 
& 44.3 $\scriptstyle\textcolor{darkgreen}{(\textbf{+3.7})}$ 
& 64.8 $\scriptstyle\textcolor{darkgreen}{(\textbf{+4.6})}$ 
& 30.2 $\scriptstyle\textcolor{darkgreen}{(\textbf{+1.2})}$ \\
& Linear Incline Weights 
& 48.5 $\scriptstyle\textcolor{darkgreen}{(\textbf{+4.3})}$ 
& 46.4 $\scriptstyle\textcolor{darkgreen}{(\textbf{+2.4})}$ 
& 31.7 $\scriptstyle\textcolor{darkgreen}{(\textbf{+1.9})}$ 
& 44.6 $\scriptstyle\textcolor{darkgreen}{(\textbf{+4.0})}$ 
& 65.0 $\scriptstyle\textcolor{darkgreen}{(\textbf{+4.8})}$ 
& 30.4 $\scriptstyle\textcolor{darkgreen}{(\textbf{+1.4})}$ \\
& Uniform Weights (Baseline GRPO) 
& 47.8 $\scriptstyle\textcolor{darkgreen}{(\textbf{+3.6})}$ 
& 45.9 $\scriptstyle\textcolor{darkgreen}{(\textbf{+1.9})}$ 
& 31.6 $\scriptstyle\textcolor{darkgreen}{(\textbf{+1.8})}$ 
& 43.7 $\scriptstyle\textcolor{darkgreen}{(\textbf{+3.1})}$ 
& 64.8 $\scriptstyle\textcolor{darkgreen}{(\textbf{+4.6})}$ 
& 29.3 $\scriptstyle\textcolor{darkgreen}{(\textbf{+0.3})}$ \\
& Inverse Parabolic Weights 
& 46.5 $\scriptstyle\textcolor{darkgreen}{(\textbf{+2.3})}$ 
& 45.1 $\scriptstyle\textcolor{darkgreen}{(\textbf{+1.1})}$ 
& 30.8 $\scriptstyle\textcolor{darkgreen}{(\textbf{+1.0})}$ 
& 42.8 $\scriptstyle\textcolor{darkgreen}{(\textbf{+2.2})}$ 
& 63.5 $\scriptstyle\textcolor{darkgreen}{(\textbf{+3.3})}$ 
& 29.1 $\scriptstyle\textcolor{darkgreen}{(\textbf{+0.1})}$ \\
 \rowcolor{PastaYellow}& \textbf{TAS (Ours)} 
& \textbf{49.1} $\scriptstyle\textcolor{darkgreen}{(\textbf{+4.9})}$ 
& \textbf{47.0} $\scriptstyle\textcolor{darkgreen}{(\textbf{+3.0})}$ 
& \textbf{32.1} $\scriptstyle\textcolor{darkgreen}{(\textbf{+2.3})}$ 
& \textbf{45.1} $\scriptstyle\textcolor{darkgreen}{(\textbf{+4.5})}$ 
& \textbf{65.6} $\scriptstyle\textcolor{darkgreen}{(\textbf{+5.4})}$ 
& \textbf{30.8} $\scriptstyle\textcolor{darkgreen}{(\textbf{+1.8})}$ \\
\bottomrule
\end{tabular}
\end{adjustbox}
\end{table*}

\paragraph{\modelname{} Resolves Vanishing Advantages.}
Figure~\ref{fig:advantages} shows that GRPO’s advantage distribution collapses to zero, confirming that groups often lack the contrast between high- and low-reward samples needed to drive learning. \modelname{} addresses this by using a stratified replay buffer to ensure reward variance. By replaying low-reward and high-reward samples within the same batch, the buffer maintains a non-zero learning signal, shifting the distribution from a zero-centered peak to a bimodal shape. This enables effective optimization by preserving meaningful reward differences, rather than stalling under uniform outcomes.

\paragraph{How efficient is \modelname{}?} 
Figure~\ref{fig:training_comparison} shows that \modelname{} reaches 0.80 accuracy in 400 iterations (1,600 unique completions), while GRPO fails to reach this performance even after 1,000 iterations (8,000 unique completions), achieving 5$\times$ greater sample efficiency (Figure~\ref{fig:training_comparison}). GRPO exhibits frequent reward collapses to zero, reflecting the vanishing advantage problem, while \modelname{} never drops below 0.3. By maintaining reward diversity within each training group, \modelname{} avoids uniform-failure scenarios that stall learning and reduces sample requirements by 80\%.

\subsection{Ablation Studies}

\paragraph{RQ1: What is the effect of each component in \modelname{}?}

To evaluate our contributions, we isolate gains relative to the on-policy GRPO baseline and evaluate on challenging video reasoning benchmarks: Video-MMMU~\cite{videommmu}, VSI-Bench~\cite{vsibench}, and Video-TT~\cite{videott}. 
As shown in Table~\ref{tab:component_ablation}, our results reveal complementary roles for the two components. The replay buffer is a primary contributor for audio-visual tasks, surpassing GRPO by $+2.4$ on OmniBench, suggesting that replaying difficult samples helps address sparse reward signals in cross-modal alignment. In contrast, TAS is particularly effective for long-horizon video reasoning, outperforming the buffer on VSI-Bench ($+1.1$ vs. $+0.7$ over baseline) and Video-TT ($+2.0$ vs.\ $+1.5$). This indicates that position-dependent weighting of planning and synthesis phases is beneficial for these tasks. Combining both components on top of GRPO yields the strongest performance (e.g., $+2.6$ on Video-MMMU), demonstrating that data diversity and structured credit assignment are complementary.

\paragraph{RQ2: What is the effect of the training curriculum and advantage shaping?}
Table~\ref{tab:curriculum_reward_ablation} validates the effectiveness of both the staged training curriculum and the TAS strategy. The curriculum shows consistent stage-wise improvements: Stage $1$ RL strengthens visual reasoning over SFT, Stage $2$ adds cross-modal alignment, and Stage $3$ refines fine-grained localization, yielding progressive gains in reasoning ability. The advantage shaping ablation compares several weighting schemes: uniform (GRPO baseline), linear decay (planning-focused), linear incline (synthesis-focused), inverse parabolic (middle-focused), and our parabolic TAS (planning and synthesis-focused). Results confirm that extreme emphasis on either phase or the middle yields inferior results, while TAS achieves the best overall performance by jointly amplifying early planning and late synthesis steps.

\begin{figure}[!t]
    \centering
    \begin{subfigure}[b]{0.49\columnwidth}
        \centering
        \includegraphics[width=\textwidth]{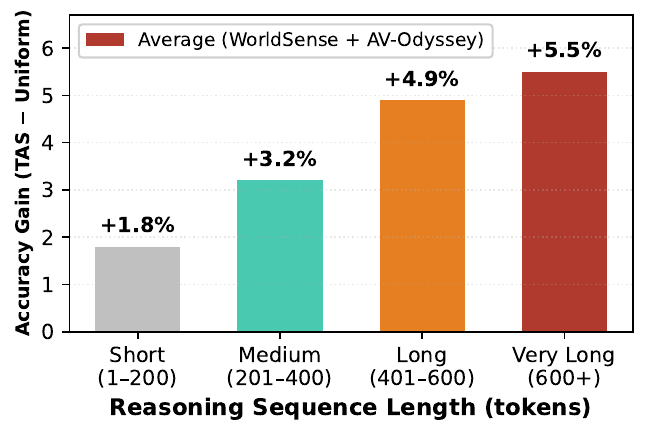}
        \caption{Audio-Visual Reasoning}
    \end{subfigure}
    \hfill
    \begin{subfigure}[b]{0.49\columnwidth}
        \centering
        \includegraphics[width=\textwidth]{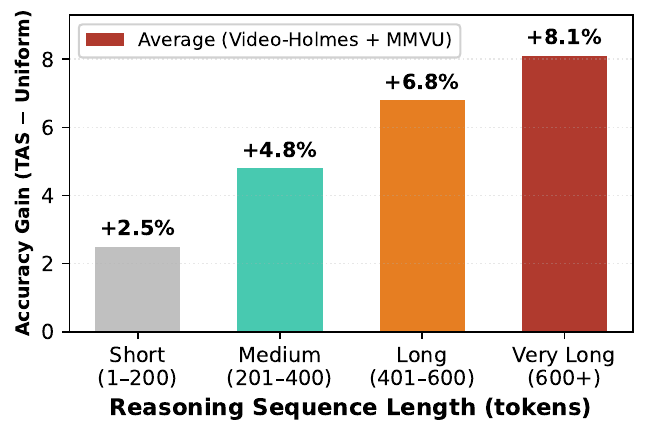}
        \caption{Video Reasoning}
    \end{subfigure}
    \caption{
        \textbf{Effect of reasoning sequence length on TAS performance.}
        In (a) audio-visual and (b) video reasoning benchmarks, 
        \textbf{TAS yields greater gains with longer reasoning sequences}.
    }
    \label{fig:tas_length}
\end{figure}

\paragraph{RQ3: Impact of TAS across Reasoning Lengths.}
As shown in Figure~\ref{fig:tas_length}, TAS consistently improves performance across reasoning lengths, with gains amplifying as sequences grow longer. 
The advantage becomes more pronounced for extended sequences ($400+$ tokens), particularly on challenging benchmarks like MMVU and Video-Holmes. 
This highlights TAS’s ability to preserve credit assignment and stabilize optimization in long-horizon reasoning, where uniform credit tends to dilute reward signals.

\section{Conclusion}
We introduce \modelname{}, an RL framework that addresses key limitations of GRPO in multimodal audio-video reasoning. Our primary finding is that jointly addressing data inefficiency and uniform credit assignment is essential for robust performance. We show that an off-policy curriculum with a stratified replay buffer outperforms standard on-policy methods, while TAS significantly enhances reasoning quality by emphasizing critical planning and synthesis phases. Looking ahead, extending \modelname{} to streaming audio-visual reasoning is a promising direction for future work.

\section*{Acknowledgments}
This research was supported by the National Eye Institute (NEI) of the National Institutes of Health (NIH) under award number R01EY034562.\ The content is solely the responsibility of the authors and does not necessarily represent the official views of the NIH. 
{
    \small
    \bibliographystyle{ieeenat_fullname}
    \bibliography{main}

@String(CVPR= {IEEE Conf. Comput. Vis. Pattern Recog.})

@String(ICCV= {Int. Conf. Comput. Vis.})

@String(ECCV= {Eur. Conf. Comput. Vis.})

@String(ICLR = {Int. Conf. Learn. Represent.})

@String(AAAI = {AAAI})

@String(CVPR  = {CVPR})

@String(ICCV  = {ICCV})

@String(ECCV  = {ECCV})

@String(ICLR  = {ICLR})

@article{avreasoner,
  title={AV-Reasoner: Improving and Benchmarking Clue-Grounded Audio-Visual Counting for MLLMs},
  author={Lu, Lidong and Chen, Guo and Li, Zhiqi and Liu, Yicheng and Lu, Tong},
  journal={arXiv preprint arXiv:2506.05328},
  year={2025}
}

@inproceedings{omnir1,
      title={Omni-R1: Reinforcement Learning for Omnimodal Reasoning via Two-System Collaboration}, 
      author={Hao Zhong and Muzhi Zhu and Zongze Du and Zheng Huang and Canyu Zhao and et al.},

      year={2025},
      booktitle={Neural Information Processing Systems (NeurIPS)}


}

@article{deepseek,
  title={Deepseek-r1: Incentivizing reasoning capability in llms via reinforcement learning},
  author={Guo, Daya and Yang, Dejian and Zhang, Haowei and Song, Junxiao and Zhang, Ruoyu and et al.},

  journal={arXiv preprint arXiv:2501.12948},
  year={2025}
}

@article{sftorrl,
      title={SFT or RL? An Early Investigation into Training R1-Like Reasoning Large Vision-Language Models}, 
      author={Hardy Chen and Haoqin Tu and Fali Wang and Hui Liu and Xianfeng Tang and et al.},

      year={2025},
    journal={Transactions on Machine Learning Research (TMLR)},
}

@InProceedings{Liu_2025_ICCV,
    author    = {Liu, Ziyu and Sun, Zeyi and Zang, Yuhang and Dong, Xiaoyi and Cao, Yuhang and et al.},

    title     = {Visual-RFT: Visual Reinforcement Fine-Tuning},
    booktitle = {International Conference on Computer Vision (ICCV)},


    year      = {2025},
}

@inproceedings{videorft,
      title={VideoRFT: Incentivizing Video Reasoning Capability in MLLMs via Reinforced Fine-Tuning}, 
      author={Qi Wang and Yanrui Yu and Ye Yuan and Rui Mao and Tianfei Zhou},
      year={2025},
      booktitle={Neural Information Processing Systems (NeurIPS)}


}

@article{repo,
  title={RePO: Replay-Enhanced Policy Optimization},
  author={Li, Siheng and Zhou, Zhanhui and Lam, Wai and Yang, Chao and Lu, Chaochao},
  journal={arXiv preprint arXiv:2506.09340},
  year={2025}
}

@inproceedings{vlrethinker,
      title={VL-Rethinker: Incentivizing Self-Reflection of Vision-Language Models with Reinforcement Learning}, 
      author={Haozhe Wang and Chao Qu and Zuming Huang and Wei Chu and Fangzhen Lin and et al.},

      year={2025},
      booktitle={Neural Information Processing Systems (NeurIPS)}


}

@inproceedings{understandingr1zero,
      title={Understanding R1-Zero-Like Training: A Critical Perspective}, 
      author={Zichen Liu and Changyu Chen and Wenjun Li and Penghui Qi and Tianyu Pang and et al.},

      year={2025},
      booktitle={Conference on Language Modeling (COLM)}


}

@inproceedings{sgrpo,
      title={S-GRPO: Early Exit via Reinforcement Learning in Reasoning Models}, 
      author={Muzhi Dai and Chenxu Yang and Qingyi Si},
      year={2025},
      booktitle={Neural Information Processing Systems (NeurIPS)}


}

@article{rlonlyinname,
  title={RL in Name Only? Analyzing the Structural Assumptions in RL post-training for LLMs},
  author={Samineni, Soumya Rani and Kalwar, Durgesh and Valmeekam, Karthik and Stechly, Kaya and Kambhampati, Subbarao},
  journal={arXiv preprint arXiv:2505.13697},
  year={2025}
}

@inproceedings{nthr,
      title={On the Effect of Negative Gradient in Group Relative Deep Reinforcement Optimization}, 
      author={Wenlong Deng and Yi Ren and Muchen Li and Danica J. Sutherland and Xiaoxiao Li and et al.},

      year={2025},
      booktitle={Neural Information Processing Systems (NeurIPS)}


}

@article{twgrpo,
  title={Reinforcing video reasoning with focused thinking},
  author={Dang, Jisheng and Wu, Jingze and Wang, Teng and Lin, Xuanhui and Zhu, Nannan and et al.},

  journal={arXiv preprint arXiv:2505.24718},
  year={2025}
}

@inproceedings{sophia,
  title={SophiaVL-R1: Reinforcing MLLMs Reasoning with Thinking Reward},
  author={Fan, Kaixuan and Feng, Kaituo and Lyu, Haoming and Zhou, Dongzhan and Yue, Xiangyu},
  booktitle={International Conference on Learning Representations (ICLR)},


  year={2026}
}

@article{videocap,
  title={VideoCap-R1: Enhancing MLLMs for Video Captioning via Structured Thinking},
  author={Meng, Desen and Huang, Rui and Dai, Zhilin and Li, Xinhao and Xu, Yifan and et al.},

  journal={arXiv preprint arXiv:2506.01725},
  year={2025}
}

@inproceedings{hintgrpo,
 author={Qihan Huang and Weilong Dai and Jinlong Liu and Wanggui He and Hao Jiang and et al.},

      title={Boosting MLLM Reasoning with Text-Debiased Hint-GRPO}, 
    booktitle = {International Conference on Computer Vision (ICCV)},


    year      = {2025},
}

@inproceedings{videor1,
      title={Video-R1: Reinforcing Video Reasoning in MLLMs}, 
      author={Kaituo Feng and Kaixiong Gong and Bohao Li and Zonghao Guo and Yibing Wang and et al.},

      year={2025},
      booktitle={Neural Information Processing Systems (NeurIPS)}


}

@article{tpo,
  title={Temporal preference optimization for long-form video understanding},
  author={Li, Rui and Wang, Xiaohan and Zhang, Yuhui and Zohar, Orr and Wang, Zeyu and et al.},

  journal={arXiv preprint arXiv:2501.13919},
  year={2025}
}

@article{kimi,
  title={Kimi-audio technical report},
  author={Ding, Ding and Ju, Zeqian and Leng, Yichong and Liu, Songxiang and Liu, Tong and et al.},

  journal={arXiv preprint arXiv:2504.18425},
  year={2025}
}

@inproceedings{avqa,
author = {Yang, Pinci and Wang, Xin and Duan, Xuguang and Chen, Hong and Hou, Runze and et al.},

title = {AVQA: A Dataset for Audio-Visual Question Answering on Videos},
year = {2022},
booktitle = {the ACM International Conference on Multimedia (MM)},


}

@inproceedings{avsbench,
  title={Audio--visual segmentation},
  author={Zhou, Jinxing and Wang, Jianyuan and Zhang, Jiayi and Sun, Weixuan and Zhang, Jing and et al.},

  booktitle={European Conference on Computer Vision (ECCV)},


  year={2022},
}

@inproceedings{omnibench,
  title={Omnibench: Towards the future of universal omni-language models},
  author={Li, Yizhi and Zhang, Ge and Ma, Yinghao and Yuan, Ruibin and Zhu, Kang and et al.},

  booktitle={Neural Information Processing Systems (NeurIPS)},


  year={2024}
}

@article{dailyomni,
  title={Daily-Omni: Towards Audio-Visual Reasoning with Temporal Alignment across Modalities},
  author={Zhou, Ziwei and Wang, Rui and Wu, Zuxuan},
  journal={arXiv preprint arXiv:2505.17862},
  year={2025}
}

@article{avodyssey,
  title={AV-Odyssey Bench: Can Your Multimodal LLMs Really Understand Audio-Visual Information?},
  author={Gong, Kaixiong and Feng, Kaituo and Li, Bohao and Wang, Yibing and Cheng, Mofan and et al.},

  journal={arXiv preprint arXiv:2412.02611},
  year={2024}
}

@inproceedings{worlsense,
  title={Worldsense: Evaluating real-world omnimodal understanding for multimodal llms},
  author={Hong, Jack and Yan, Shilin and Cai, Jiayin and Jiang, Xiaolong and Hu, Yao and et al.},

  booktitle={International Conference on Learning Representations (ICLR)},


  year={2026}
}

@article{humanomni,
  title={Humanomni: A large vision-speech language model for human-centric video understanding},
  author={Zhao, Jiaxing and Yang, Qize and Peng, Yixing and Bai, Detao and Yao, Shimin and et al.},

  journal={arXiv preprint arXiv:2501.15111},
  year={2025}
}

@inproceedings{mvbench,
  title={Mvbench: A comprehensive multi-modal video understanding benchmark},
  author={Li, Kunchang and Wang, Yali and He, Yinan and Li, Yizhuo and Wang, Yi and et al.},

  booktitle={Conference on Computer Vision and Pattern Recognition (CVPR)},


  year={2024},
}

@inproceedings{videomme,
  title={Video-mme: The first-ever comprehensive evaluation benchmark of multi-modal llms in video analysis},
  author={Fu, Chaoyou and Dai, Yuhan and Luo, Yongdong and Li, Lei and Ren, Shuhuai and et al.},

  booktitle={Conference on Computer Vision and Pattern Recognition (CVPR)},


  year={2025}
}

@inproceedings{lvbench,
      title={LVBench: An Extreme Long Video Understanding Benchmark}, 
      author={Weihan Wang and Zehai He and Wenyi Hong and Yean Cheng and Xiaohan Zhang and et al.},

      year={2025},
      booktitle={International Conference on Computer Vision (ICCV)}


}

@article{videoholmes,
  title={Video-Holmes: Can MLLM Think Like Holmes for Complex Video Reasoning?},
  author={Cheng, Junhao and Ge, Yuying and Wang, Teng and Ge, Yixiao and Liao, Jing and et al.},

  journal={arXiv preprint arXiv:2505.21374},
  year={2025}
}

@InProceedings{mmvu,
    author    = {Zhao, Yilun and Zhang, Haowei and Xie, Lujing and Hu, Tongyan and Gan, Guo and et al.},

    title     = {MMVU: Measuring Expert-Level Multi-Discipline Video Understanding},
    booktitle = {Conference on Computer Vision and Pattern Recognition (CVPR)},


    year      = {2025},
}

@inproceedings{tomato,
  title={TOMATO: Assessing Visual Temporal Reasoning Capabilities in Multimodal Foundation Models},
  author={Shangguan, Ziyao and Li, Chuhan and Ding, Yuxuan and Zheng, Yanan and Zhao, Yilun and et al.},

  booktitle={International Conference on Learning Representations (ICLR)},


  year={2025}
}

@article{qwen25omni,
  title={Qwen2.5-omni technical report},
  author={Xu, Jin and Guo, Zhifang and He, Jinzheng and Hu, Hangrui and He, Ting and et al.},

  journal={arXiv preprint arXiv:2503.20215},
  year={2025}
}

@article{echoink,
  title={Echoink-r1: Exploring audio-visual reasoning in multimodal llms via reinforcement learning},
  author={Xing, Zhenghao and Hu, Xiaowei and Fu, Chi-Wing and Wang, Wenhai and Dai, Jifeng and et al.},

  journal={arXiv preprint arXiv:2505.04623},
  year={2025}
}

@article{ola,
  title={Ola: Pushing the frontiers of omni-modal language model},
  author={Liu, Zuyan and Dong, Yuhao and Wang, Jiahui and Liu, Ziwei and Hu, Winston and et al.},

  journal={arXiv preprint arXiv:2502.04328},
  year={2025}
}

@article{qwen25vl,
  title={Qwen2.5-vl technical report},
  author={Bai, Shuai and Chen, Keqin and Liu, Xuejing and Wang, Jialin and Ge, Wenbin and et al.},

  journal={arXiv preprint arXiv:2502.13923},
  year={2025}
}

@inproceedings{swift,
  title={Swift: a scalable lightweight infrastructure for fine-tuning},
  author={Zhao, Yuze and Huang, Jintao and Hu, Jinghan and Wang, Xingjun and Mao, Yunlin and et al.},

  booktitle={the AAAI Conference on Artificial Intelligence (AAAI)},


  year={2025}
}

@article{internvl3,
  title={Internvl3: Exploring advanced training and test-time recipes for open-source multimodal models},
  author={Zhu, Jinguo and Wang, Weiyun and Chen, Zhe and Liu, Zhaoyang and Ye, Shenglong and et al.},

  journal={arXiv preprint arXiv:2504.10479},
  year={2025}
}

@inproceedings{scalingrl,
      title={Scaling RL to Long Videos}, 
      author={Yukang Chen and Wei Huang and Baifeng Shi and Qinghao Hu and Hanrong Ye and et al.},

      booktitle={Neural Information Processing Systems (NeurIPS)},


      year={2025},
}

@inproceedings{videorts,
      title={Video-RTS: Rethinking Reinforcement Learning and Test-Time Scaling for Efficient and Enhanced Video Reasoning}, 
      author={Ziyang Wang and Jaehong Yoon and Shoubin Yu and Md Mohaiminul Islam and Gedas Bertasius and et al.},

      year={2025},
      booktitle={Empirical Methods in Natural Language Processing (EMNLP)}


}

@inproceedings{papo,
  title={Perception-aware policy optimization for multimodal reasoning},
  author={Wang, Zhenhailong and Guo, Xuehang and Stoica, Sofia and Xu, Haiyang and Wang, Hongru and et al.},

  booktitle={International Conference on Learning Representations (ICLR)},


  year={2026}
}

@inproceedings{ave,
  title={Audio-visual event localization in unconstrained videos},
  author={Tian, Yapeng and Shi, Jing and Li, Bochen and Duan, Zhiyao and Xu, Chenliang},
  booktitle={the European Conference on Computer Vision (ECCV)},


  year={2018}
}

@inproceedings{srpo,
      title={SRPO: Enhancing Multimodal LLM Reasoning via Reflection-Aware Reinforcement Learning}, 
      author={Zhongwei Wan and Zhihao Dou and Che Liu and Yu Zhang and Dongfei Cui and et al.},

      year={2025},
      booktitle={Neural Information Processing Systems (NeurIPS)},


}

@inproceedings{videopasta,
      title={VideoPASTA: 7K Preference Pairs That Matter for Video-LLM Alignment}, 
      author={Yogesh Kulkarni and Pooyan Fazli},
      year={2025},
      booktitle={Empirical Methods in Natural Language Processing (EMNLP)}


}

@inproceedings{videosavi,
      title={VideoSAVi: Self-Aligned Video Language Models without Human Supervision}, 
      author={Yogesh Kulkarni and Pooyan Fazli},
      year={2025},
     booktitle={Conference on Language Modeling (COLM)}


}

@inproceedings{rrpo,
      title={Self-alignment of Large Video Language Models with Refined Regularized Preference Optimization}, 
      author={Pritam Sarkar and Ali Etemad},
      year={2025},
      booktitle={Neural Information Processing Systems (NeurIPS)}


}

@inproceedings{dpo,
  title={Direct preference optimization: Your language model is secretly a reward model},
  author={Rafailov, Rafael and Sharma, Archit and Mitchell, Eric and Manning, Christopher D and Ermon, Stefano and et al.},

  booktitle={Neural Information Processing Systems (NeurIPS)},


  year={2023}
}

@inproceedings{fastslow,
      title={Fast-Slow Thinking for Large Vision-Language Model Reasoning}, 
      author={Wenyi Xiao and Leilei Gan and Weilong Dai and Wanggui He and Ziwei Huang and et al.},

      year={2025},
      booktitle={Neural Information Processing Systems (NeurIPS)}


}

@article{wei2025advancing,
  title={Advancing Multimodal Reasoning via Reinforcement Learning with Cold Start},
  author={Wei, Lai and Li, Yuting and Zheng, Kaipeng and Wang, Chen and Wang, Yue and et al.},

  journal={arXiv preprint arXiv:2505.22334},
  year={2025}
}

@article{li2025reinforcement,
  title={Reinforcement Learning Tuning for VideoLLMs: Reward Design and Data Efficiency},
  author={Li, Hongyu and Han, Songhao and Liao, Yue and Luo, Junfeng and Gao, Jialin and et al.},

  journal={arXiv preprint arXiv:2506.01908},
  year={2025}
}

@article{pixelthink,
  title={PixelThink: Towards Efficient Chain-of-Pixel Reasoning},
  author={Wang, Song and Fang, Gongfan and Kong, Lingdong and Li, Xiangtai and Xu, Jianyun and et al.},

  journal={arXiv preprint arXiv:2505.23727},
  year={2025}
}

@inproceedings{grit,
      title={GRIT: Teaching MLLMs to Think with Images}, 
      author={Yue Fan and Xuehai He and Diji Yang and Kaizhi Zheng and Ching-Chen Kuo and et al.},

      year={2025},
      booktitle={Neural Information Processing Systems (NeurIPS)}


}

@inproceedings{grounded,
      title={Grounded Reinforcement Learning for Visual Reasoning}, 
      author={Gabriel Sarch and Snigdha Saha and Naitik Khandelwal and Ayush Jain and Michael J. Tarr and et al.},

      year={2025},
      booktitle={Neural Information Processing Systems (NeurIPS)}


}

@article{veripo,
  title={VerIPO: Cultivating Long Reasoning in Video-LLMs via Verifier-Gudied Iterative Policy Optimization},
  author={Li, Yunxin and Chen, Xinyu and Li, Zitao and Liu, Zhenyu and Wang, Longyue and et al.},

  journal={arXiv preprint arXiv:2505.19000},
  year={2025}
}

@inproceedings{bimba,
  title={Bimba: Selective-scan compression for long-range video question answering},
  author={Islam, Md Mohaiminul and Nagarajan, Tushar and Wang, Huiyu and Bertasius, Gedas and Torresani, Lorenzo},
  booktitle={Conference on Computer Vision and Pattern Recognition (CVPR)},


  year={2025}
}

@inproceedings{salmonn,
  title={video-SALMONN-o1: Reasoning-enhanced Audio-visual Large Language Model},
  author={Sun, Guangzhi and Yang, Yudong and Zhuang, Jimin and Tang, Changli and Li, Yixuan and et al.},

  booktitle={International Conference on Machine Learning (ICML)},


  year={2025}
}

@article{vita,
  title={Vita: Towards open-source interactive omni multimodal llm},
  author={Fu, Chaoyou and Lin, Haojia and Long, Zuwei and Shen, Yunhang and Dai, Yuhang and et al.},

  journal={arXiv preprint arXiv:2408.05211},
  year={2024}
}

@inproceedings{empowering,
  title={Empowering llms with pseudo-untrimmed videos for audio-visual temporal understanding},
  author={Tang, Yunlong and Shimada, Daiki and Bi, Jing and Feng, Mingqian and Hua, Hang and et al.},

  booktitle={the AAAI Conference on Artificial Intelligence (AAAI)},


  year={2025}
}

@article{videollama2,
  title={Videollama 2: Advancing spatial-temporal modeling and audio understanding in video-llms},
  author={Cheng, Zesen and Leng, Sicong and Zhang, Hang and Xin, Yifei and Li, Xin and et al.},

  journal={arXiv preprint arXiv:2406.07476},
  year={2024}
}

@inproceedings{cat,
  title={Cat: Enhancing multimodal large language model to answer questions in dynamic audio-visual scenarios},
  author={Ye, Qilang and Yu, Zitong and Shao, Rui and Xie, Xinyu and Torr, Philip and et al.},

  booktitle={European Conference on Computer Vision (ECCV)},


  year={2024},
}

@inproceedings{verl,
  title={Hybridflow: A flexible and efficient rlhf framework},
  author={Sheng, Guangming and Zhang, Chi and Ye, Zilingfeng and Wu, Xibin and Zhang, Wang and et al.},

  booktitle={the Twentieth European Conference on Computer Systems (EuroSys)},


  year={2025}
}

@inproceedings{hu2021lora,
  title={Lora: Low-rank adaptation of large language models.},
  author={Hu, Edward J and Shen, Yelong and Wallis, Phillip and Allen-Zhu, Zeyuan and Li, Yuanzhi and et al.},

  booktitle={International Conference on Learning Representations (ICLR)},


  year={2022}
}

@article{feng2025don,
  title={Don't Waste Mistakes: Leveraging Negative RL-Groups via Confidence Reweighting},
  author={Feng, Yunzhen and Jain, Parag and Hartshorn, Anthony and Duan, Yaqi and Kempe, Julia},
  journal={arXiv preprint arXiv:2510.08696},
  year={2025}
}

@inproceedings{vidhalluc,
    author    = {Li, Chaoyu and Im, Eun Woo and Fazli, Pooyan},
    title     = {VidHalluc: Evaluating Temporal Hallucinations in Multimodal Large Language Models for Video Understanding},
    booktitle = {Conference on Computer Vision and Pattern Recognition (CVPR)},


    year      = {2025},
}

@inproceedings{yu2025dapo,
  title={Dapo: An open-source llm reinforcement learning system at scale},
  author={Yu, Qiying and Zhang, Zheng and Zhu, Ruofei and Yuan, Yufeng and Zuo, Xiaochen and et al.},

  booktitle={Neural Information Processing Systems (NeurIPS)},


  year={2025}
}

@inproceedings{xiao2023efficient,
  title={Efficient streaming language models with attention sinks},
  author={Xiao, Guangxuan and Tian, Yuandong and Chen, Beidi and Han, Song and Lewis, Mike},
  booktitle={International Conference on Learning Representations (ICLR)},


  year={2024}
}

@article{kulkarni2025egovita,
  title={EgoVITA: Learning to Plan and Verify for Egocentric Video Reasoning},
  author={Kulkarni, Yogesh and Fazli, Pooyan},
  journal={arXiv preprint arXiv:2511.18242},
  year={2025}
}

@inproceedings{schaul2015prioritized,
  title={Prioritized experience replay},
  author={Schaul, Tom and Quan, John and Antonoglou, Ioannis and Silver, David},
  booktitle={International Conference on Learning Representations (ICLR)},


  year={2016}
}

@article{videommmu,
  title={Video-mmmu: Evaluating knowledge acquisition from multi-discipline professional videos},
  author={Hu, Kairui and Wu, Penghao and Pu, Fanyi and Xiao, Wang and Zhang, Yuanhan and et al.},

  journal={arXiv preprint arXiv:2501.13826},
  year={2025}
}

@inproceedings{vsibench,
  title={Thinking in space: How multimodal large language models see, remember, and recall spaces},
  author={Yang, Jihan and Yang, Shusheng and Gupta, Anjali W and Han, Rilyn and Fei-Fei, Li and et al.},

  booktitle={Conference on Computer Vision and Pattern Recognition (CVPR)},


  year={2025}
}

@inproceedings{videott,
  title={Towards video thinking test: A holistic benchmark for advanced video reasoning and understanding},
  author={Zhang, Yuanhan and Chew, Yunice and Dong, Yuhao and Leo, Aria and Hu, Bo and et al.},

  booktitle={International Conference on Computer Vision (ICCV)},


  year={2025}
}

@inproceedings{visionr1,
  title={Vision-r1: Incentivizing reasoning capability in multimodal large language models},
  author={Huang, Wenxuan and Jia, Bohan and Zhai, Zijie and Cao, Shaosheng and Ye, Zheyu and et al.},

  booktitle={International Conference on Learning Representations (ICLR)},


  year={2026}
}

@inproceedings{openvl,
  title={Openvlthinker: Complex vision-language reasoning via iterative sft-rl cycles},
  author={Deng, Yihe and Bansal, Hritik and Yin, Fan and Peng, Nanyun and Wang, Wei and et al.},

  booktitle={Neural Information Processing Systems (NeurIPS)},


  year={2025}
}

@inproceedings{mathverse,
  title={Mathverse: Does your multi-modal llm truly see the diagrams in visual math problems?},
  author={Zhang, Renrui and Jiang, Dongzhi and Zhang, Yichi and Lin, Haokun and Guo, Ziyu and et al.},

  booktitle={European Conference on Computer Vision (ECCV)},


  year={2024},
}

@inproceedings{mathvista,
  title={Mathvista: Evaluating mathematical reasoning of foundation models in visual contexts},
  author={Lu, Pan and Bansal, Hritik and Xia, Tony and Liu, Jiacheng and Li, Chunyuan and et al.},

  booktitle={International Conference on Learning Representations (ICLR)},


  year={2024}
}

@article{mmk12,
  title={Mm-eureka: Exploring the frontiers of multimodal reasoning with rule-based reinforcement learning},
  author={Meng, Fanqing and Du, Lingxiao and Liu, Zongkai and Zhou, Zhixiang and Lu, Quanfeng and et al.},

  journal={arXiv preprint arXiv:2503.07365},
  year={2025}
}

@inproceedings{r1ov,
  title={R1-onevision: Advancing generalized multimodal reasoning through cross-modal formalization},
  author={Yang, Yi and He, Xiaoxuan and Pan, Hongkun and Jiang, Xiyan and Deng, Yan and et al.},

  booktitle={International Conference on Computer Vision (ICCV)},


  year={2025}
}

@article{scivideobench,
  title={Scivideobench: Benchmarking scientific video reasoning in large multimodal models},
  author={Deng, Andong and Yang, Taojiannan and Yu, Shoubin and Spencer, Lincoln and Bansal, Mohit and et al.},

  journal={arXiv preprint arXiv:2510.08559},
  year={2025}
}
}
\clearpage
\setcounter{page}{1}

\maketitlesupplementary

\section{GRPO for Multimodal Reasoning}
Recent works increasingly use RL methods, particularly GRPO~\cite{fastslow, srpo, papo}, introducing new challenges related to training efficiency and stability. Wei et al. propose a ``cold start" strategy via Supervised Fine-Tuning (SFT) before RL~\cite{wei2025advancing}, while other approaches improve data efficiency through direct RL with test-time scaling
~\cite{videorts}. Concurrent advances enhance GRPO through curriculum learning~\cite{avreasoner}, variance-aware data selection~\cite{li2025reinforcement}, hint-based guidance for difficult samples~\cite{hintgrpo}, and combining GRPO and DPO~\cite{veripo}. Key limitations remain, including the ``overthinking'' issue, addressed by regulating reasoning length~\cite{pixelthink, fastslow}, and perception errors mitigated via grounded reasoning approaches~\cite{grit, grounded} and new RL objectives such as PAPO’s Implicit Perception Loss~\cite{papo}.

Audio-visual reasoning that incorporates audio often involves trade-offs between temporal coverage and spatial resolution. These challenges have been addressed through two-system architectures~\cite{omnir1}, explicit contextual rewards~\cite{humanomni}, and task-specific improvements~\cite{avreasoner, echoink}. \modelname{} introduces a unified framework that integrates a stratified replay buffer for curriculum learning with Temporal Advantage Shaping (TAS) for credit assignment. This approach improves reasoning and grounding within a single model, eliminating the need for two-system architectures~\cite{omnir1} or explicit visual coordinate generation~\cite{grounded}.

\section{Ablation Studies}

\begin{figure}[t!] 
\centering  
\includegraphics[width=\columnwidth]{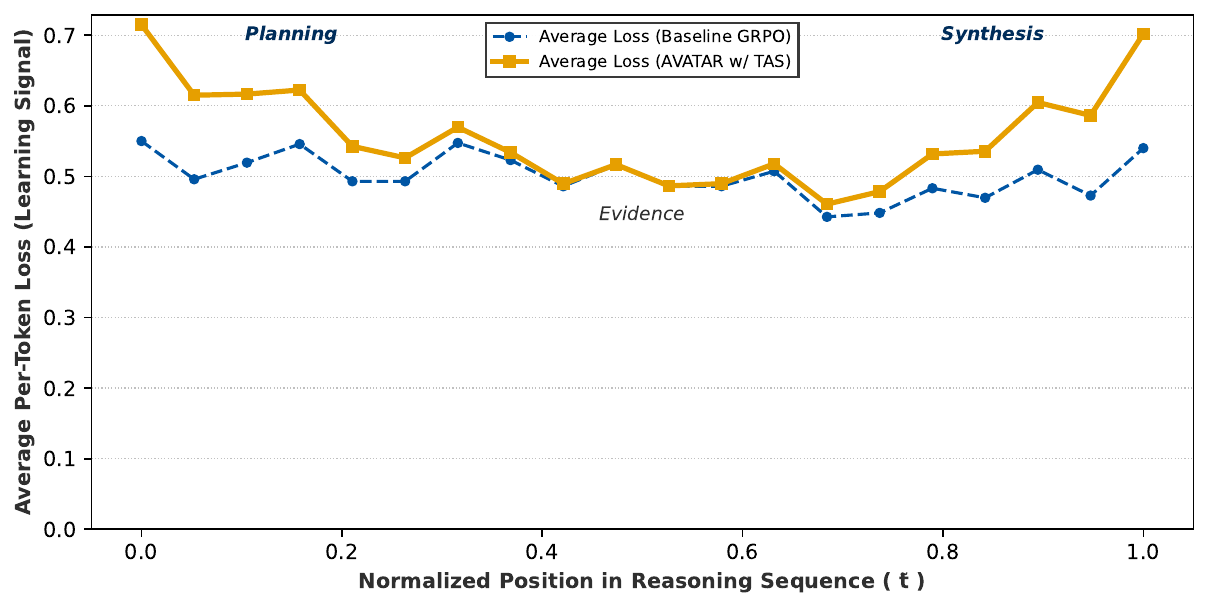} 
\caption{\textbf{TAS Empirically Focuses the Learning Signal.} 
We plot the average per-token loss (the learning signal) vs. normalized position across the Video-Holmes~\cite{videoholmes} benchmark. The \textbf{Baseline GRPO} (\textcolor{AcademicBlue}{blue dashed line}) shows a flat, unfocused signal, indicative of \textbf{uniform credit assignment}. \textbf{AVATAR w/ TAS} (\textcolor{ContrastingOrange}{orange solid line}) successfully focuses the learning signal, amplifying it on the critical \textit{planning} and \textit{synthesis} phases. }
 \label{fig:tas_loss_viz} 
\end{figure}

\paragraph{RQ1: Does TAS empirically focus the learning signal in practice?} We validate that TAS successfully adopts this parabolic-shape in practice. We perform an analysis on the Video-Holmes~\cite{videoholmes} benchmark, calculating the average per-token loss (the learning signal) against its normalized position. Figure~\ref{fig:tas_loss_viz} shows the result. The baseline GRPO (\textcolor{AcademicBlue}{blue dashed line}) exhibits a relatively flat loss distribution. This is the empirical evidence of \textit{uniform credit assignment}: the learning signal is diluted by being applied unfocused and equally to all tokens, regardless of their importance. In contrast, \modelname{} w/ TAS (\textcolor{ContrastingOrange}{orange solid line}) demonstrates a clear parabolic-shaped loss profile. The shaped advantage $A_{i,t}^{TAS}$ amplifies the learning signal at the critical \textit{planning} ($\tilde{t} \in [0, 0.2]$) and \textit{synthesis} ($\tilde{t} \in [0.8, 1.0]$) phases. This proves TAS is a functioning mechanism: it forces the optimizer to learn most aggressively from errors in the most important parts of the reasoning chain.
\begin{table}[!t]
\centering
\caption{\textbf{Ablation on key reasoning tasks from the WorldSense benchmark}. Halluc: Hallucination, TP: Temporal Prediction, AC: Audio Change}
\label{tab:world_ablation}
\begin{adjustbox}{width=\columnwidth,center}
\renewcommand{\arraystretch}{1.2}
\fontsize{7pt}{8pt}\selectfont
\setlength{\tabcolsep}{2mm}
\begin{tabular}{l ccc}
\toprule
\textbf{Config} & \textbf{Halluc} & \textbf{TP} & \textbf{AC} \\
\midrule
Baseline GRPO & 35.6 & 53.6 & 32.5 \\
\midrule
AVATAR w/o Off-Policy 
& 40.1 $\scriptstyle\textcolor{darkgreen}{(\textbf{+4.5})}$ 
& 54.8 $\scriptstyle\textcolor{darkgreen}{(\textbf{+1.2})}$ 
& 34.1 $\scriptstyle\textcolor{darkgreen}{(\textbf{+1.6})}$ \\
AVATAR w/o TAS 
& 48.9 $\scriptstyle\textcolor{darkgreen}{(\textbf{+13.3})}$ 
& 56.1 $\scriptstyle\textcolor{darkgreen}{(\textbf{+2.5})}$ 
& 35.8 $\scriptstyle\textcolor{darkgreen}{(\textbf{+3.3})}$ \\
\midrule
\rowcolor{PastaYellow}
\textbf{AVATAR (Full)} 
& \textbf{51.2} $\scriptstyle\textcolor{darkgreen}{(\textbf{+15.6})}$ 
& \textbf{57.2} $\scriptstyle\textcolor{darkgreen}{(\textbf{+3.6})}$ 
& \textbf{37.3} $\scriptstyle\textcolor{darkgreen}{(\textbf{+4.8})}$ \\
\bottomrule
\end{tabular}
\end{adjustbox}
\end{table}
\begin{figure*}[t]
 \centering
 \begin{subfigure}[b]{0.32\linewidth}
   \centering
   \includegraphics[width=\linewidth]{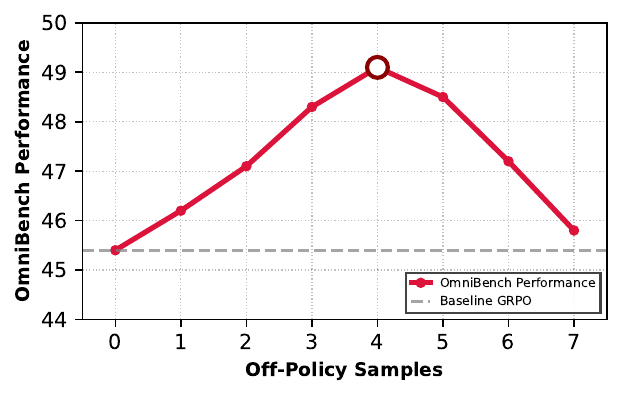}
   \caption{OmniBench}
   \label{fig:omnibench}
 \end{subfigure}
 \hfill
 \begin{subfigure}[b]{0.32\linewidth}
   \centering
   \includegraphics[width=\linewidth]{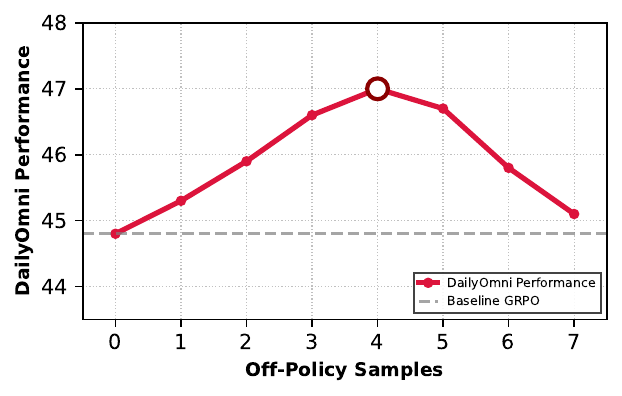}
   \caption{DailyOmni}
   \label{fig:dailyomni}
 \end{subfigure}
 \hfill
 \begin{subfigure}[b]{0.32\linewidth}
   \centering
   \includegraphics[width=\linewidth]{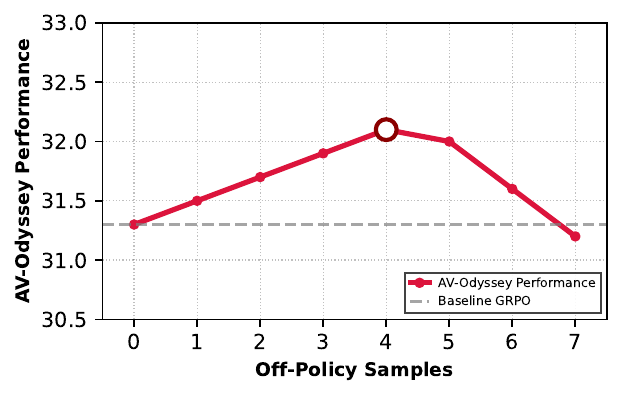}
   \caption{AV-Odyssey}
   \label{fig:avodyssey}
 \end{subfigure}
 
 \vspace{0.1cm}
 \begin{subfigure}[b]{0.32\linewidth}
   \centering
   \includegraphics[width=\linewidth]{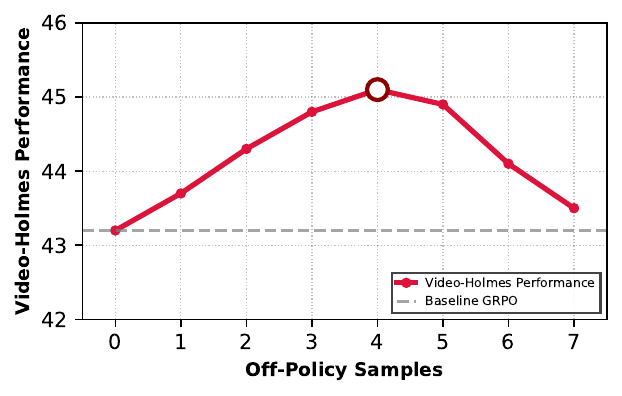}
   \caption{Video-Holmes}
   \label{fig:videoholmes}
 \end{subfigure}
 \hfill
 \begin{subfigure}[b]{0.32\linewidth}
   \centering
   \includegraphics[width=\linewidth]{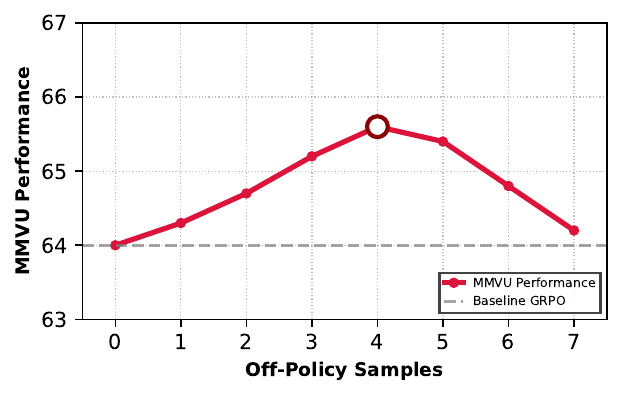}
   \caption{MMVU}
   \label{fig:mmvu}
 \end{subfigure}
 \hfill
 \begin{subfigure}[b]{0.32\linewidth}
   \centering
   \includegraphics[width=\linewidth]{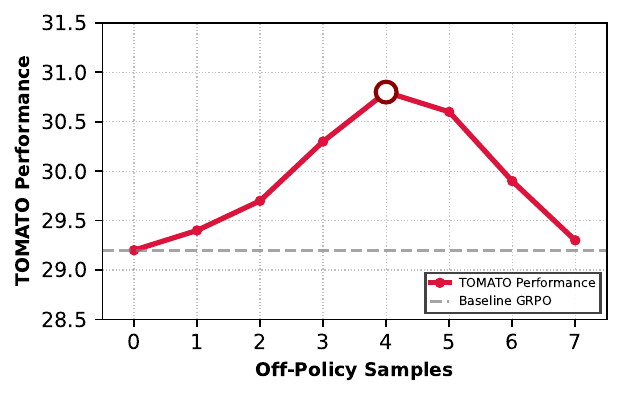}
   \caption{TOMATO}
   \label{fig:tomato}
 \end{subfigure}
 
\caption{\textbf{Performance vs. on-policy/off-policy sample ratio across six benchmarks}. \textbf{AVATAR achieves optimal performance with $4$-$4$ split} ($4$ on-policy, $4$ off-policy samples), demonstrating that balanced mixing prevents both policy drift from excessive off-policy data and sample inefficiency from pure on-policy training.}
 \label{fig:onoff_policy_analysis}
\end{figure*}

\paragraph{RQ2: How do \modelname{}'s components impact audio-visual reasoning?}

The impact is evident on reasoning tasks from the WorldSense benchmark (Table~\ref{tab:world_ablation}). The highest improvement is on the \textit{Hallucination} task, a gain primarily driven by our off-policy architecture, which builds a more robust and grounded model by forcing repeated engagement with difficult samples from the hard tier. The effect of TAS is most clear on tasks requiring temporal logic, such as \textit{Temporal Prediction}, which involves forecasting future events, and \textit{Audio Change}, which requires identifying discrete shifts in the audio stream. The U-shaped weighting of TAS is uniquely suited for these problems, by modifying credit for initial tokens, it forces accurate grounding of the video’s initial state, while modifying credit for final tokens rewards correct synthesis of how that state evolved over time.

\begin{table}[!t]
\centering
\caption{\textbf{Ablation comparing the impact of \textbf{SFT} and \textbf{AVATAR} integration}. Models trained with SFT achieve \textbf{stronger cross-modal grounding} and \textbf{reward alignment}, while removing SFT leads to significant degradation in both GRPO and AVATAR.}

\label{tab:av_ablation_simple}
\begin{adjustbox}{width=\columnwidth,center}
\renewcommand{\arraystretch}{1.2}
\fontsize{7pt}{8pt}\selectfont
\setlength{\tabcolsep}{3mm}
\begin{tabular}{l ccc}
\toprule
\textbf{Setting} & \multicolumn{3}{c}{\textbf{Audio-Visual Benchmarks}} \\
\cmidrule(lr){2-4}
& \textbf{OmniBench} & \textbf{DailyOmni} & \textbf{AV-Odyssey} \\
\midrule
Qwen2.5 Omni & 44.2 & 44.0 & 29.8 \\
\midrule
GRPO (w/o SFT) & 41.2 $\scriptstyle\textcolor{red}{(\textbf{-3.0})}$ & 40.8 $\scriptstyle\textcolor{red}{(\textbf{-3.2})}$ & 27.1 $\scriptstyle\textcolor{red}{(\textbf{-2.7})}$ \\
GRPO (w/ SFT) & 46.0 $\scriptstyle\textcolor{darkgreen}{(\textbf{+1.8})}$ & 45.5 $\scriptstyle\textcolor{darkgreen}{(\textbf{+1.5})}$ & 30.2 $\scriptstyle\textcolor{darkgreen}{(\textbf{+0.4})}$ \\
\midrule
\textbf{AVATAR} (w/o SFT) & 43.8 $\scriptstyle\textcolor{red}{(\textbf{-0.4})}$ & 43.5 $\scriptstyle\textcolor{red}{(\textbf{-0.5})}$ & 28.9 $\scriptstyle\textcolor{red}{(\textbf{-0.9})}$ \\
\rowcolor{PastaYellow}
\textbf{AVATAR} (w/ SFT) & \textbf{49.1} $\scriptstyle\textcolor{darkgreen}{(\textbf{+4.9})}$ & \textbf{47.0} $\scriptstyle\textcolor{darkgreen}{(\textbf{+3.0})}$ & \textbf{32.1} $\scriptstyle\textcolor{darkgreen}{(\textbf{+2.3})}$ \\
\bottomrule
\end{tabular}
\end{adjustbox}
\end{table}

\paragraph{RQ3: What is the impact of the on-policy/off-policy sample ratio?}
To determine the optimal balance between on-policy and off-policy data, we analyze performance across different sample ratios within a fixed group size of eight (Figure~\ref{fig:onoff_policy_analysis}). We observe a consistent trend across all benchmarks: performance peaks when using a balanced $4$-$4$ split ($4$ on-policy, $4$ off-policy samples). Using fewer off-policy samples ($0$-$3$) fails to fully mitigate the data inefficiency of a purely on-policy approach. On the other hand, over-reliance on off-policy samples ($5$-$7$) degrades performance due to instability, as the policy drifts too far from the older, stale data in the replay buffer.
The $4$-$4$ split therefore represents an empirical sweet spot for \modelname{}, maximizing the data efficiency gains of our off-policy architecture while maintaining the training stability necessary for effective learning.

\paragraph{RQ4: Effect of SFT on Learning Stability.}
Table~\ref{tab:av_ablation_simple} shows that incorporating SFT before GRPO provides essential structural grounding that improves convergence and generalization. 
Without SFT, models lack cross-modal alignment, resulting in unstable rewards and degraded performance across all benchmarks. 
Initializing GRPO from an SFT-trained policy yields more stable learning and moderate gains, while AVATAR achieves the highest improvements. 
Hence, SFT establishes a reliable foundation for stable advantage estimation and TAS-based reward shaping, enabling GRPO to refine rather than relearn multimodal reasoning.

\paragraph{RQ5: \modelname{} is Generalizes Beyond Video.} 

We train Qwen2.5-VL-7B~\cite{qwen25vl} on the Vision-R1-RL dataset~\cite{visionr1} using both standard GRPO and \modelname{}, and evaluate on four multimodal math reasoning benchmarks: MathVerse~\cite{mathverse}, MathVista~\cite{mathvista}, MMK12~\cite{mmk12}, and R1-OneVision-Bench~\cite{r1ov}. As shown in Table~\ref{tab:math_results}, \modelname{} generalizes effectively beyond video, outperforming reasoning models such as OpenVLThinker~\cite{openvl} and VLAAThinker~\cite{sftorrl}, with gains of $+9.9$ on MMK12 and $+3.6$ on MathVerse. We further observe a distinct ablation pattern compared to audio-visual tasks, where Hinting and TAS emerge as the primary contributors to performance. This difference arises because math reasoning relies on multi-step dependencies, rather than the sparse modality signals of video alignment. In this setting, a single intermediate error can break the entire solution, making TAS (which prevents attention drift) and Hinting (which provides strategic guidance for approaching complex problems) critical for success.

\begin{table}[t]
  \centering
  \caption{\textbf{Evaluation on multimodal math reasoning.} \modelname{} demonstrates strong task-agnostic capabilities by achieving strong results across all four benchmarks.}
  \label{tab:math_results}
  \begin{adjustbox}{max width=\linewidth,center}
    \renewcommand{\arraystretch}{1.05}
    \fontsize{6pt}{7pt}\selectfont
    \setlength{\tabcolsep}{1.2mm}
    \begin{tabular}{lcccc}
      \toprule
      \textbf{Model} & \textbf{MathVerse} & \textbf{MathVista} & \textbf{MMK12} & \textbf{R1-OV Bench} \\

      \rowcolor{gray!15}\multicolumn{5}{c}{\textit{State-of-the-Art Methods}} \\
      OpenVLThinker-7B~\cite{openvl} & 45.8 & 70.0 & 53.5 & 34.7 \\
      VLAAThinker-VL-7B~\cite{sftorrl} & 48.2 & 68.0 & 51.7 & 38.4 \\
      \midrule
      \rowcolor{gray!15}\multicolumn{5}{c}{\textit{Baseline}} \\
      Qwen2.5-VL-7B & 46.0 & 67.1 & 47.9 & 34.9 \\
      + GRPO & 48.1 & 68.7 & 56.3 & 36.8 \\
      \rowcolor{PastaYellow}
      + \textbf{\modelwithlogo} & \textbf{49.6} $\scriptstyle\textcolor{darkgreen}{(\textbf{+3.6})}$ & 69.8 $\scriptstyle\textcolor{darkgreen}{(\textbf{+2.7})}$ & \textbf{57.8} $\scriptstyle\textcolor{darkgreen}{(\textbf{+9.9})}$ & \textbf{38.9} $\scriptstyle\textcolor{darkgreen}{(\textbf{+4.0})}$ \\
      \midrule
      \rowcolor{gray!15}\multicolumn{5}{c}{\textit{Ablation Studies}} \\
      \quad w/o Replay Buffer (on-policy) & 49.2 $\scriptstyle\textcolor{darkgreen}{(\textbf{+3.2})}$ & \textbf{70.4} $\scriptstyle\textcolor{darkgreen}{(\textbf{+3.3})}$ & 57.0 $\scriptstyle\textcolor{darkgreen}{(\textbf{+9.1})}$ & 38.5 $\scriptstyle\textcolor{darkgreen}{(\textbf{+3.6})}$ \\
      \quad w/o TAS (uniform credit) & 48.4 $\scriptstyle\textcolor{darkgreen}{(\textbf{+2.4})}$ & 69.2 $\scriptstyle\textcolor{darkgreen}{(\textbf{+2.1})}$ & 56.2 $\scriptstyle\textcolor{darkgreen}{(\textbf{+8.3})}$ & 37.5 $\scriptstyle\textcolor{darkgreen}{(\textbf{+2.6})}$ \\
      \quad w/o Hinting & 48.1 $\scriptstyle\textcolor{darkgreen}{(\textbf{+2.1})}$ & 68.8 $\scriptstyle\textcolor{darkgreen}{(\textbf{+1.7})}$ & 55.6 $\scriptstyle\textcolor{darkgreen}{(\textbf{+7.7})}$ & 37.0 $\scriptstyle\textcolor{darkgreen}{(\textbf{+2.1})}$ \\
      \bottomrule
    \end{tabular}
  \end{adjustbox}
\end{table}
\begin{figure}[t]
  \centering
  \includegraphics[width=\columnwidth]{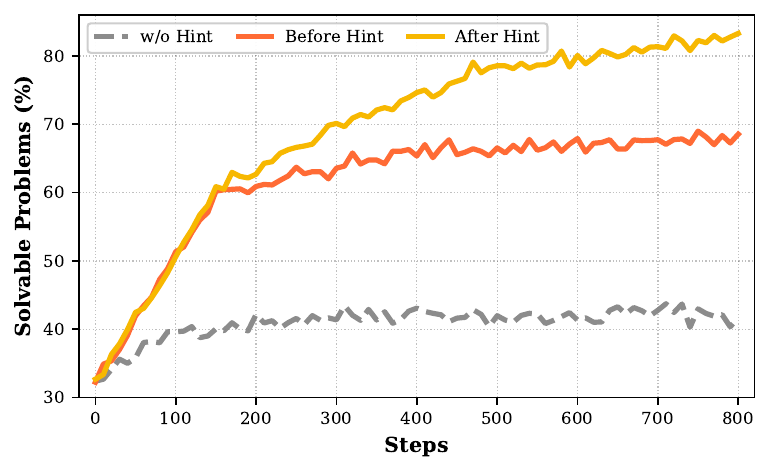}
  \caption{\textbf{Hinting unlocks solvability.} While GRPO (grey) stagnates, \modelname{} (orange) learns steadily.}
  \label{fig:hint_impact}
\end{figure}

\paragraph{RQ6: Hinting Unlocks the Hard Tail of the Distribution.} 

Figure~\ref{fig:hint_impact} quantifies the impact of the hinting mechanism on problem solvability during training. The baseline GRPO (grey) stagnates early ($\approx\!40\%$), failing to learn from complex failures. \modelname{} without hinting (orange) improves solvability to $\approx\!68\%$ via off-policy replay but eventually plateaus. Enabling hinting (yellow) initiates a second phase of growth, increasing solvability to over $\mathbf{80\%}$. This demonstrates that hints effectively unlock the hardest tail of the distribution, namely, problems that otherwise yield zero reward, by converting them into informative learning signals.

\paragraph{RQ7: Efficacy on Reasoning-Heavy Video Benchmarks.}
To validate \modelname{} on tasks demanding complex temporal logic and domain-specific knowledge, we evaluate the Qwen2.5-Omni-7B model across four reasoning-intensive datasets: Video-MMMU~\cite{videommmu}, VSI-Bench~\cite{vsibench}, SciVideoBench~\cite{scivideobench}, and Video-TT~\cite{videott} (Table~\ref{tab:video_reasoning}). While standard GRPO yields only marginal improvements over the base model due to uniform credit assignment failing on long-horizon tasks, \modelname{} consistently delivers robust performance. It achieves absolute gains of $+2.6$ on Video-MMMU and $+2.2$ on SciVideoBench over the base model, outperforming GRPO across all metrics. These results confirm that our stratified replay and temporal advantage shaping successfully sustain exploration on challenging reasoning paths where standard on-policy RL collapses.
\begin{table}[!t]
\centering
\caption{\textbf{\modelname{} on reasoning-heavy video benchmarks.}}
\label{tab:video_reasoning}
\begin{adjustbox}{width=\columnwidth,center}
\renewcommand{\arraystretch}{1.2}
\fontsize{7pt}{8pt}\selectfont
\setlength{\tabcolsep}{2mm}
\begin{tabular}{lcccc}
\toprule
\textbf{Model} & \textbf{Video-MMMU} & \textbf{VSI-Bench} & \textbf{SciVideoBench} & \textbf{Video-TT} \\
\midrule
\rowcolor{gray!15}\multicolumn{5}{c}{\textit{Baseline}} \\
Qwen2.5-Omni-7B & 46.8 & 25.4 & 29.2 & 41.8  \\
\quad + GRPO & 48.1 & 25.9 & 30.3 & 43.0 \\
\rowcolor{PastaYellow}
\quad + \textbf{\modelwithlogo} & \textbf{49.4} $\scriptstyle\textcolor{darkgreen}{(\textbf{+2.6)}}$ & \textbf{26.8} $\scriptstyle\textcolor{darkgreen}{(\textbf{+1.4})}$ & \textbf{31.4} $\scriptstyle\textcolor{darkgreen}{(\textbf{+2.2})}$ & \textbf{44.2} $\scriptstyle\textcolor{darkgreen}{(\textbf{+2.4})}$  \\
\bottomrule
\end{tabular}
\end{adjustbox}
\end{table}

\begin{figure}[!t]
    \centering
    \begin{subfigure}[b]{0.9\columnwidth}
        \centering
        \includegraphics[width=\textwidth]{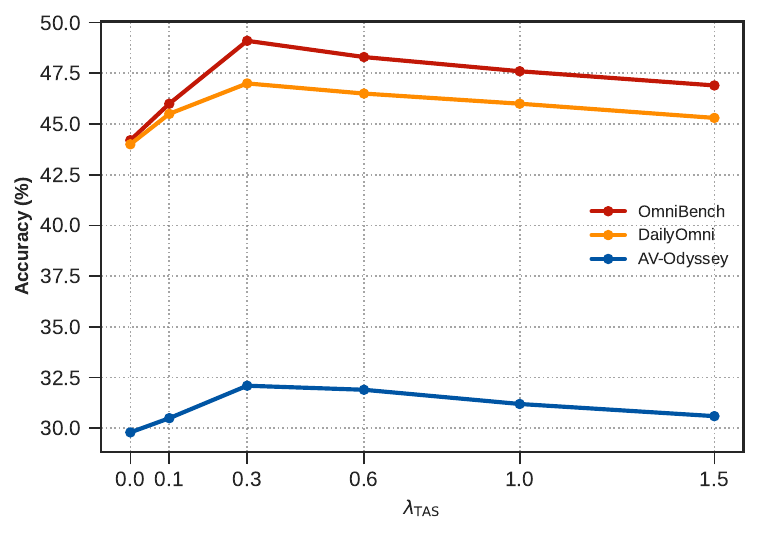}
        \caption{Audio-Visual Reasoning}
    \end{subfigure}
    \vspace{2mm} 
    \begin{subfigure}[b]{0.9\columnwidth}
        \centering
        \includegraphics[width=\textwidth]{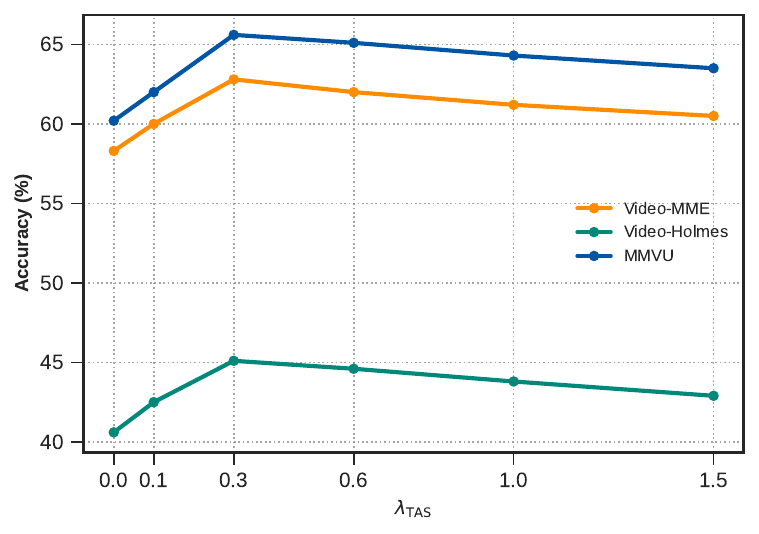}
        \caption{Video Reasoning}
    \end{subfigure}
    \caption{\textbf{Performance variation with TAS weighting $(\lambda_{\mathrm{TAS}})$}.
    Both (a) \textbf{Audio-Visual} and (b) \textbf{Video Reasoning} benchmarks peak around \textbf{$\lambda_{\mathrm{TAS}}{=}0.3$}, showing improved stability and reward shaping before slight decline at higher values.}
    \label{fig:tas_sweep}
\end{figure}

\paragraph{RQ8: Impact of $\lambda_{\mathrm{TAS}}$ on Learning Stability.}
As illustrated in Figure~\ref{fig:tas_sweep}, model performance exhibits a clear three-phase trend with respect to $\lambda_{\mathrm{TAS}}$.  At low values ($\lambda_{\mathrm{TAS}}{<}0.1$), the reward shaping effect is weak, leading to slower convergence and unstable gains. 
As $\lambda_{\mathrm{TAS}}$ increases to a moderate level ($\lambda_{\mathrm{TAS}}{\approx}0.3$), performance peaks across both audio-visual and video reasoning benchmarks, reflecting a balanced trade-off between stability and exploration. 
However, at higher values ($\lambda_{\mathrm{TAS}}{>}1.0$), the model becomes over-regularized, reducing adaptability and leading to gradual accuracy decline.

\paragraph{RQ9: How do isolating rewards impact performance?} 
\begin{table*}[!t]
\centering
\caption{\textbf{Ablation studies of the reward strategy across three training stages}. Stage $1$ uses $R_{\mathrm{acc}}$ and $R_{\mathrm{format}}$ for basic accuracy and format compliance. Stage $2$ adds $R_{\mathrm{self}}$ (majority vote consensus) to enable more consistent learning. Stage $3$ incorporates $R_{\mathrm{judge}}$ (stepwise VLM evaluation) alongside previous rewards, providing more detailed reasoning feedback. VCRS maintains stable advantage baselines via moving averages, preventing vanishing advantages. Each stage yields consistent performance gains, with the largest improvements on reasoning-heavy benchmarks (Video-Holmes: $+4.5$, MMVU: $+5.4$), where structured feedback is most beneficial.}
\label{tab:reward_ablation}
\begin{adjustbox}{width=\textwidth,center}
\renewcommand{\arraystretch}{1.2}
\fontsize{7pt}{8pt}\selectfont
\setlength{\tabcolsep}{2mm}
\begin{tabular}{llcccccc}
\toprule
\multicolumn{2}{c}{\textbf{Ablation Configuration}} & \multicolumn{3}{c}{\textbf{Audio-Visual Benchmarks}} & \multicolumn{3}{c}{\textbf{Video Reasoning Benchmarks}} \\
\cmidrule(r){1-2} \cmidrule(lr){3-5} \cmidrule(l){6-8}
\textbf{Group} & \textbf{Setting} & \textbf{OmniBench} & \textbf{DailyOmni} & \textbf{AV-Odyssey} & \textbf{Video-Holmes} & \textbf{MMVU} & \textbf{TOMATO} \\
\midrule
& Baseline Qwen2.5 Omni & 44.2 & 44.0 & 29.8 & 40.6 & 60.2 & 29.0 \\
\midrule
\multirow{4}{*}{\textit{Reward Suite}} 
& $R_{\mathrm{acc}} + R_{\mathrm{format}}$ 
& 46.8 $\scriptstyle\textcolor{darkgreen}{(\textbf{+2.6})}$ 
& 45.2 $\scriptstyle\textcolor{darkgreen}{(\textbf{+1.2})}$ 
& 30.1 $\scriptstyle\textcolor{darkgreen}{(\textbf{+0.3})}$ 
& 42.3 $\scriptstyle\textcolor{darkgreen}{(\textbf{+1.7})}$ 
& 62.1 $\scriptstyle\textcolor{darkgreen}{(\textbf{+1.9})}$ 
& 28.9 $\scriptstyle\textcolor{darkgreen}{(\textbf{-0.1})}$ \\
& $R_{\mathrm{acc}} + R_{\mathrm{format}} + R_{\mathrm{self}}$ 
& 47.5 $\scriptstyle\textcolor{darkgreen}{(\textbf{+3.3})}$ 
& 45.8 $\scriptstyle\textcolor{darkgreen}{(\textbf{+1.8})}$ 
& 30.7 $\scriptstyle\textcolor{darkgreen}{(\textbf{+0.9})}$ 
& 43.1 $\scriptstyle\textcolor{darkgreen}{(\textbf{+2.5})}$ 
& 63.4 $\scriptstyle\textcolor{darkgreen}{(\textbf{+3.2})}$ 
& 29.4 $\scriptstyle\textcolor{darkgreen}{(\textbf{+0.4})}$ \\
& $R_{\mathrm{acc}} + R_{\mathrm{format}} + R_{\mathrm{self}} + R_{\mathrm{judge}}$ 
& 48.3 $\scriptstyle\textcolor{darkgreen}{(\textbf{+4.1})}$ 
& 46.4 $\scriptstyle\textcolor{darkgreen}{(\textbf{+2.4})}$ 
& 31.4 $\scriptstyle\textcolor{darkgreen}{(\textbf{+1.6})}$ 
& 44.2 $\scriptstyle\textcolor{darkgreen}{(\textbf{+3.6})}$ 
& 64.7 $\scriptstyle\textcolor{darkgreen}{(\textbf{+4.5})}$ 
& 30.1 $\scriptstyle\textcolor{darkgreen}{(\textbf{+1.1})}$ \\
\rowcolor{PastaYellow}
& $R_{\mathrm{acc}} + R_{\mathrm{format}} + R_{\mathrm{self}} + R_{\mathrm{judge}} + \mathrm{VCRS}$ 
& \textbf{49.1} $\scriptstyle\textcolor{darkgreen}{(\textbf{+4.9})}$ 
& \textbf{47.0} $\scriptstyle\textcolor{darkgreen}{(\textbf{+3.0})}$ 
& \textbf{32.1} $\scriptstyle\textcolor{darkgreen}{(\textbf{+2.3})}$ 
& \textbf{45.1} $\scriptstyle\textcolor{darkgreen}{(\textbf{+4.5})}$ 
& \textbf{65.6} $\scriptstyle\textcolor{darkgreen}{(\textbf{+5.4})}$ 
& \textbf{30.8} $\scriptstyle\textcolor{darkgreen}{(\textbf{+1.8})}$ \\
\bottomrule
\end{tabular}
\end{adjustbox}
\end{table*}
Table~\ref{tab:reward_ablation} shows the cumulative impact of each reward component in \modelname{}'s training pipeline. The baseline, which includes only accuracy and format rewards, offers limited learning signals, especially on complex reasoning tasks such as Video-Holmes and MMVU, where binary feedback proves insufficient. Introducing self-rewarding ($R_{\mathrm{self}}$) in Stage $2$ enables consensus-based learning from the model’s own outputs, yielding moderate improvements on audio-visual tasks that benefit from internal consistency signals (e.g., $+0.7$ on OmniBench, $+1.3$ on MMVU). Adding the stepwise judge reward ($R_{\mathrm{judge}}$) in Stage $3$ results in larger gains by assessing intermediate reasoning steps, particularly enhancing performance on fine-grained localization benchmarks (e.g., $+0.8$ on OmniBench, $+1.3$ on MMVU). Finally, incorporating VCRS further stabilizes training by maintaining non-zero advantage baselines via moving averages, allowing the full reward setup to achieve the highest observed performance (e.g., $+0.8$ on OmniBench, $+0.9$ on MMVU).

\paragraph{RQ10: How do training dynamics differ across benchmarks?}

\begin{figure*}[!t]
 \centering
 \begin{subfigure}[b]{0.30\linewidth}
   \centering
   \includegraphics[width=\linewidth]{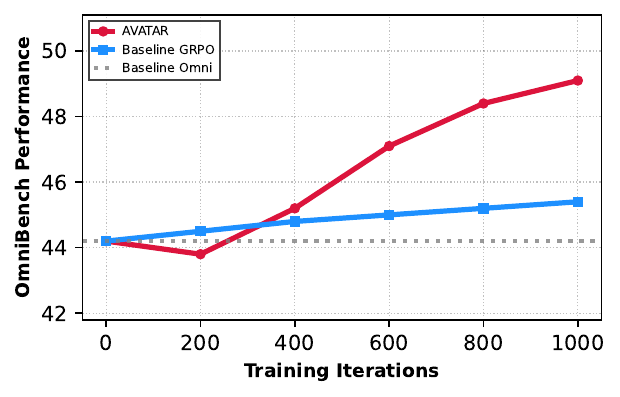}
   \caption{OmniBench}
   \label{fig:omnibench_training}
 \end{subfigure}
 \hfill
 \begin{subfigure}[b]{0.30\linewidth}
   \centering
   \includegraphics[width=\linewidth]{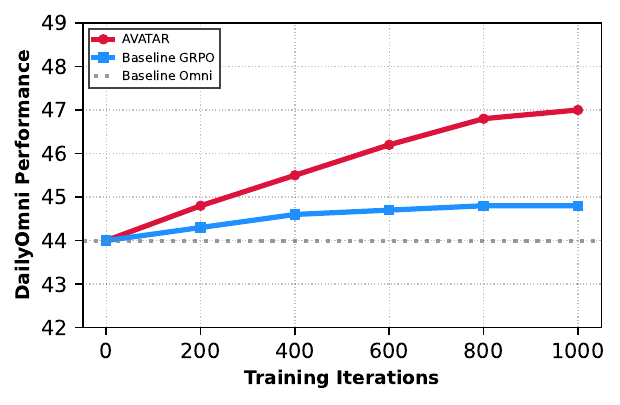}
   \caption{DailyOmni}
   \label{fig:dailyomni_training}
 \end{subfigure}
 \hfill
 \begin{subfigure}[b]{0.30\linewidth}
   \centering
   \includegraphics[width=\linewidth]{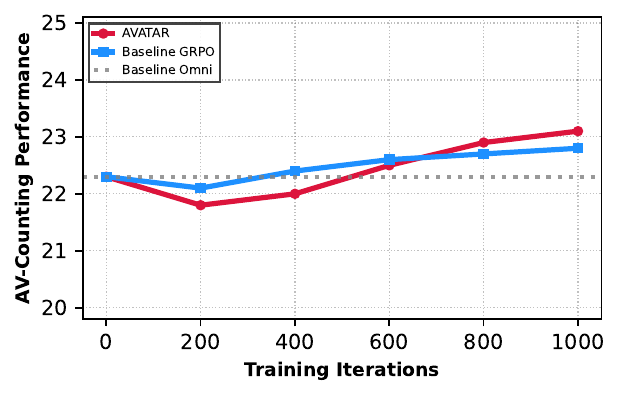}
   \caption{AV-Counting}
   \label{fig:avcounting_training}
 \end{subfigure}
 
 \vspace{0.1cm}
 \begin{subfigure}[b]{0.30\linewidth}
   \centering
   \includegraphics[width=\linewidth]{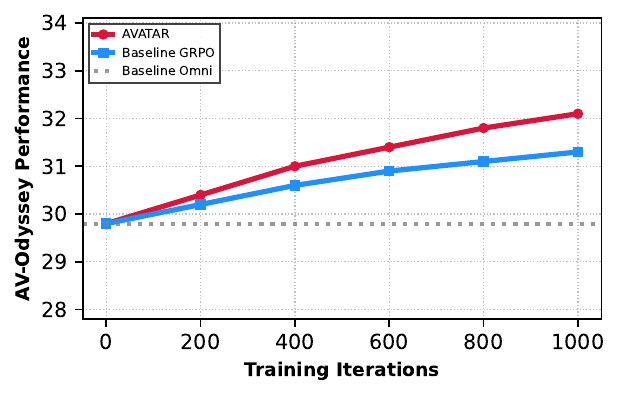}
   \caption{AV-Odyssey}
   \label{fig:avodyssey_training}
 \end{subfigure}
 \hfill
 \begin{subfigure}[b]{0.30\linewidth}
   \centering
   \includegraphics[width=\linewidth]{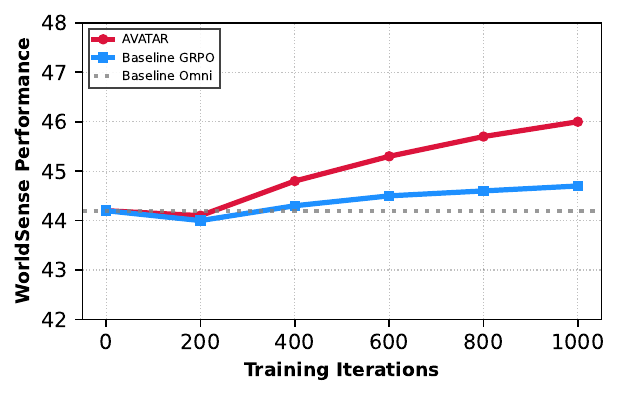}
   \caption{WorldSense}
   \label{fig:worldsense_training}
 \end{subfigure}
 \hfill
 \begin{subfigure}[b]{0.30\linewidth}
   \centering
   \includegraphics[width=\linewidth]{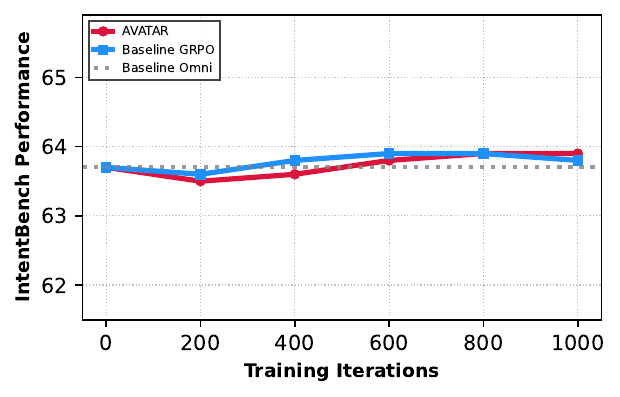}
   \caption{IntentBench}
   \label{fig:intentbench_training}
 \end{subfigure}
 
 \vspace{0.1cm}
 \begin{subfigure}[b]{0.30\linewidth}
   \centering
   \includegraphics[width=\linewidth]{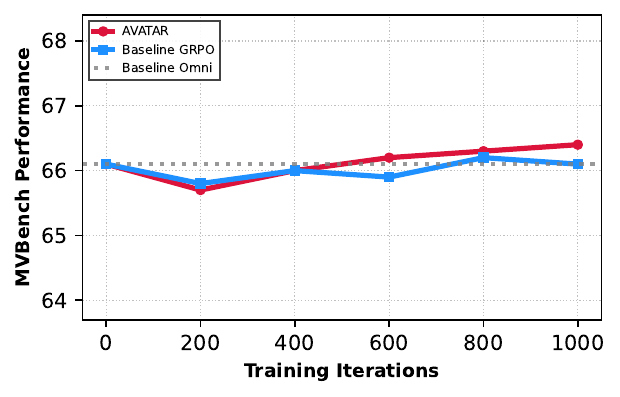}
   \caption{MVBench}
   \label{fig:mvbench_training}
 \end{subfigure}
 \hfill
 \begin{subfigure}[b]{0.30\linewidth}
   \centering
   \includegraphics[width=\linewidth]{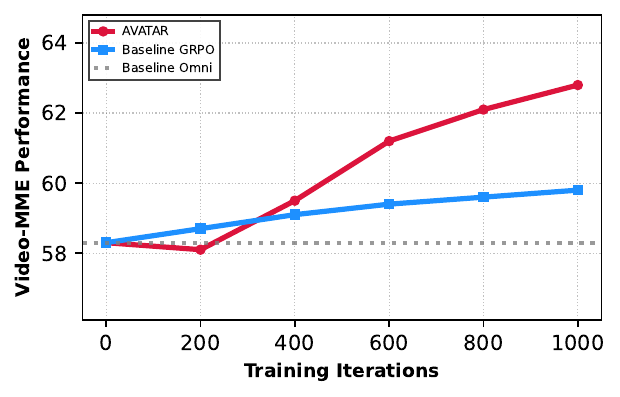}
   \caption{Video-MME}
   \label{fig:videomme_training}
 \end{subfigure}
 \hfill
 \begin{subfigure}[b]{0.30\linewidth}
   \centering
   \includegraphics[width=\linewidth]{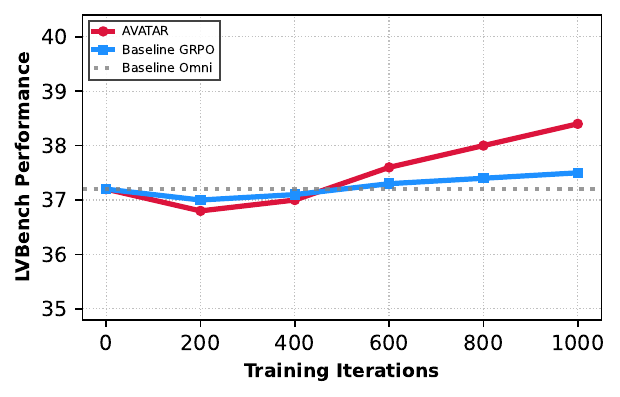}
   \caption{LVBench}
   \label{fig:lvbench_training}
 \end{subfigure}
 
 \vspace{0.1cm}
 \begin{subfigure}[b]{0.30\linewidth}
   \centering
   \includegraphics[width=\linewidth]{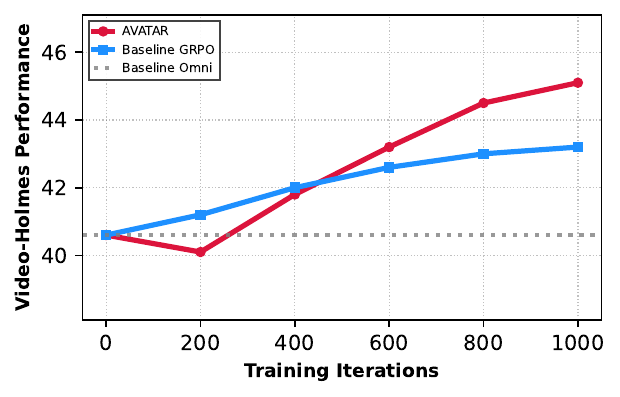}
   \caption{Video-Holmes}
   \label{fig:videoholmes_training}
 \end{subfigure}
 \hfill
 \begin{subfigure}[b]{0.30\linewidth}
   \centering
   \includegraphics[width=\linewidth]{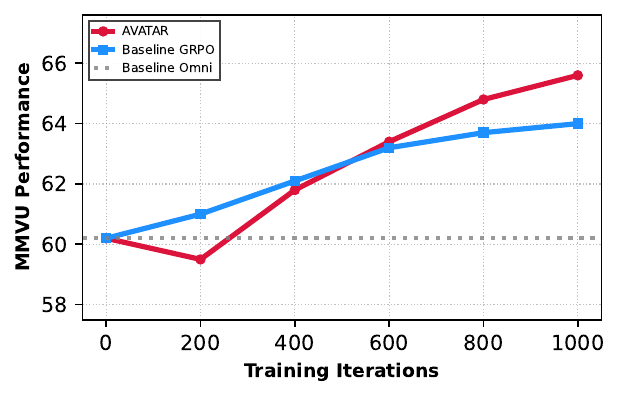}
   \caption{MMVU}
   \label{fig:mmvu_training}
 \end{subfigure}
 \hfill
 \begin{subfigure}[b]{0.30\linewidth}
   \centering
   \includegraphics[width=\linewidth]{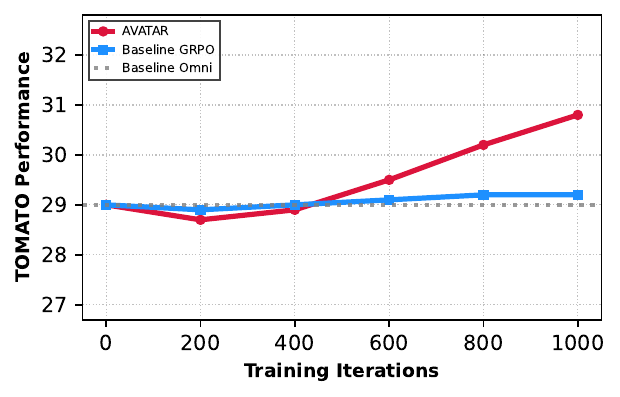}
   \caption{TOMATO}
   \label{fig:tomato_training}
 \end{subfigure}
 
\caption{\textbf{Training curves across audio-visual and video reasoning benchmarks}. \textbf{AVATAR demonstrates superior sample efficiency and final performance}, particularly on challenging reasoning tasks. AVATAR's initial dip, followed by a strong recovery, validates the effectiveness of our off-policy architecture and TAS for credit assignment.}
 \label{fig:training_curves}
\end{figure*}
The training curves in Figure~\ref{fig:training_curves} reveal distinct learning patterns that directly validate our core contributions. Baseline GRPO shows a consistent decline across all benchmarks due to the vanishing advantage problem: when encountering difficult samples where all responses receive similar rewards, the learning signal collapses to zero, stalling further learning. As a result, GRPO quickly plateaus and fails to move beyond solving only the simplest queries.

\noindent In contrast, \modelname{} shows a \textit{dip-and-recovery pattern} that reflects the interplay of its key components. 
The initial drop in performance, especially on challenging benchmarks like OmniBench, Video-Holmes, and MMVU, occurs when the stratified replay buffer transitions from easy to hard samples, exposing the model to its most difficult failures. This performance dip would persist without TAS; however, the U-shaped temporal weighting in TAS provides focused learning signals that extract meaningful gradients from these difficult examples. By emphasizing the planning and synthesis phases that uniform credit assignment would dilute, TAS enables recovery and subsequent improvement. The model’s rebound and improved final performance confirm that our off-policy architecture transforms early failures into learning opportunities through repeated exposure, while TAS ensures that each replayed experience yields maximum value. These gains emerge within just $1000$ iterations, underscoring \modelname{}’s superior sample efficiency and training effectiveness.

\paragraph{RQ11: Does AVATAR generalize to smaller models?} To evaluate the generalizability of our approach across different model scales, we conduct experiments with a $3B$ parameter version of the base Qwen2.5 Omni~\cite{qwen25omni}  model. Tables~\ref{tab:av_3b_results} and~\ref{tab:video_3b_results} demonstrate that \modelname{} consistently outperforms both the baseline and on-policy GRPO across all benchmarks, achieving improvements of $+3.4$ on OmniBench, $+2.9$ on Video-Holmes, and $+2.8$ on MMVU. These results validate that \modelname{}’s core components, the stratified replay buffer and TAS, effectively enhance multimodal reasoning capabilities regardless of language model scale. The consistent performance patterns across different model sizes further confirm that our off-policy architecture is broadly applicable, with particularly strong benefits observed on complex reasoning tasks, where \modelname{}’s targeted learning signals are most impactful in guiding model improvement.

\begin{table*}[!t]
\centering
\caption{\textbf{Audio-visual benchmarks: $3B$ model comparison}. \modelname{} 3B consistently outperforms both the baseline $3B$ Omni and baseline GRPO across all benchmarks, achieving improvements of $+3.4$ on OmniBench and $+2.6$ on AV-Odyssey. The results demonstrate that our off-policy architecture and TAS effectively enhance multimodal reasoning capabilities even with smaller model scales, validating the broad applicability of our approach. Performance improvements over baseline 3B Omni are shown in green.}
\label{tab:av_3b_results}
\begin{adjustbox}{width=\textwidth,center}
\renewcommand{\arraystretch}{1.2}
\fontsize{7pt}{8pt}\selectfont
\setlength{\tabcolsep}{2mm}
\begin{tabular}{lcccccc}
\toprule
& \multicolumn{6}{c}{\textbf{Audio-Visual Benchmarks}} \\
\midrule
\textbf{Model} & \textbf{OmniBench} & \textbf{DailyOmni} & \textbf{AV-Counting} & \textbf{AV-Odyssey} & \textbf{WorldSense} & \textbf{IntentBench} \\
\midrule
Baseline $3B$ Omni & 42.4 & 42.9 & 19.7 & 27.2 & 35.2 & 59.7 \\
Baseline GRPO 
& 43.1 $\scriptstyle\textcolor{darkgreen}{(\textbf{+0.7})}$ 
& 43.4 $\scriptstyle\textcolor{darkgreen}{(\textbf{+0.5})}$ 
& 20.0 $\scriptstyle\textcolor{darkgreen}{(\textbf{+0.3})}$ 
& 28.5 $\scriptstyle\textcolor{darkgreen}{(\textbf{+1.3})}$ 
& 35.8 $\scriptstyle\textcolor{darkgreen}{(\textbf{+0.6})}$ 
& 59.9 $\scriptstyle\textcolor{darkgreen}{(\textbf{+0.2})}$ \\
\rowcolor{PastaYellow}
\textbf{\modelname{} $3B$} 
& \textbf{45.8} $\scriptstyle\textcolor{darkgreen}{(\textbf{+3.4})}$ 
& \textbf{44.7} $\scriptstyle\textcolor{darkgreen}{(\textbf{+1.8})}$ 
& \textbf{20.9} $\scriptstyle\textcolor{darkgreen}{(\textbf{+1.2})}$ 
& \textbf{29.8} $\scriptstyle\textcolor{darkgreen}{(\textbf{+2.6})}$ 
& \textbf{37.1} $\scriptstyle\textcolor{darkgreen}{(\textbf{+1.9})}$ 
& \textbf{60.5} $\scriptstyle\textcolor{darkgreen}{(\textbf{+0.8})}$ \\
\bottomrule
\end{tabular}
\end{adjustbox}
\end{table*}
\begin{table*}[!t]
\centering
\caption{\textbf{Video understanding and reasoning benchmarks: $3B$ model comparison}. \modelname{} $3B$ demonstrates superior performance across both general video understanding and complex reasoning tasks, with particularly strong gains on reasoning-heavy benchmarks like Video-Holmes ($+2.9$) and MMVU ($+2.8$). The consistent improvements over baseline GRPO highlight how our stratified replay buffer and TAS enhance learning efficiency even with reduced model capacity, confirming that our framework’s benefits scale effectively across different model sizes. Performance improvements over baseline 3B Omni are shown in green.}
\label{tab:video_3b_results}
\begin{adjustbox}{width=\textwidth,center}
\renewcommand{\arraystretch}{1.2}
\fontsize{7pt}{8pt}\selectfont
\setlength{\tabcolsep}{2mm}
\begin{tabular}{l|ccc|ccc}
\toprule
& \multicolumn{3}{c|}{\textbf{General Video Understanding}} & \multicolumn{3}{c}{\textbf{Video Reasoning}} \\
\cmidrule(r){2-4} \cmidrule(l){5-7}
\textbf{Model} & \textbf{MVBench} & \textbf{Video-MME} & \textbf{LVBench} & \textbf{Video-Holmes} & \textbf{MMVU} & \textbf{TOMATO} \\
\midrule
Baseline $3B$ Omni & 59.2 & 57.6 & 34.3 & 39.2 & 62.0 & 27.0 \\
Baseline GRPO 
& 59.8 $\scriptstyle\textcolor{darkgreen}{(\textbf{+0.6})}$ 
& 58.1 $\scriptstyle\textcolor{darkgreen}{(\textbf{+0.5})}$ 
& 34.7 $\scriptstyle\textcolor{darkgreen}{(\textbf{+0.4})}$ 
& 40.8 $\scriptstyle\textcolor{darkgreen}{(\textbf{+1.6})}$ 
& 63.2 $\scriptstyle\textcolor{darkgreen}{(\textbf{+1.2})}$ 
& 27.3 $\scriptstyle\textcolor{darkgreen}{(\textbf{+0.3})}$ \\
\rowcolor{PastaYellow}
\textbf{\modelname{} $3B$} 
& \textbf{61.2} $\scriptstyle\textcolor{darkgreen}{(\textbf{+2.0})}$ 
& \textbf{59.9} $\scriptstyle\textcolor{darkgreen}{(\textbf{+2.3})}$ 
& \textbf{35.8} $\scriptstyle\textcolor{darkgreen}{(\textbf{+1.5})}$ 
& \textbf{42.1} $\scriptstyle\textcolor{darkgreen}{(\textbf{+2.9})}$ 
& \textbf{64.8} $\scriptstyle\textcolor{darkgreen}{(\textbf{+2.8})}$ 
& \textbf{28.4} $\scriptstyle\textcolor{darkgreen}{(\textbf{+1.4})}$ \\
\bottomrule
\end{tabular}
\end{adjustbox}
\end{table*}

\subsection{Prompt Templates}
To support \modelname{}’s training pipeline, we use two prompt templates for different components of our framework. The \textit{hint generation prompt} (Figure~\ref{fig:hint-prompt}) is used by our stratified replay buffer’s hint mechanism when the policy becomes stuck in local optima on challenging samples. This prompt instructs a teacher model to generate strategic hints without revealing the final answer, helping the policy escape stagnation while maintaining the learning challenge. The \textit{audio-visual localization judge prompt} (Figure~\ref{fig:judge_prompt_stage3}) helps our stepwise reasoning judge in Stage $3$ training by providing granular evaluation criteria across four dimensions: grounding of audio cues, identification of visual objects, spatial localization accuracy, and caption correctness.  This offers detailed feedback on cross-modal reasoning quality rather than binary success/failure signals.

\subsection{Qualitative Analysis}
\begin{figure*}[!t]
\centering
\includegraphics[width=0.9\textwidth]{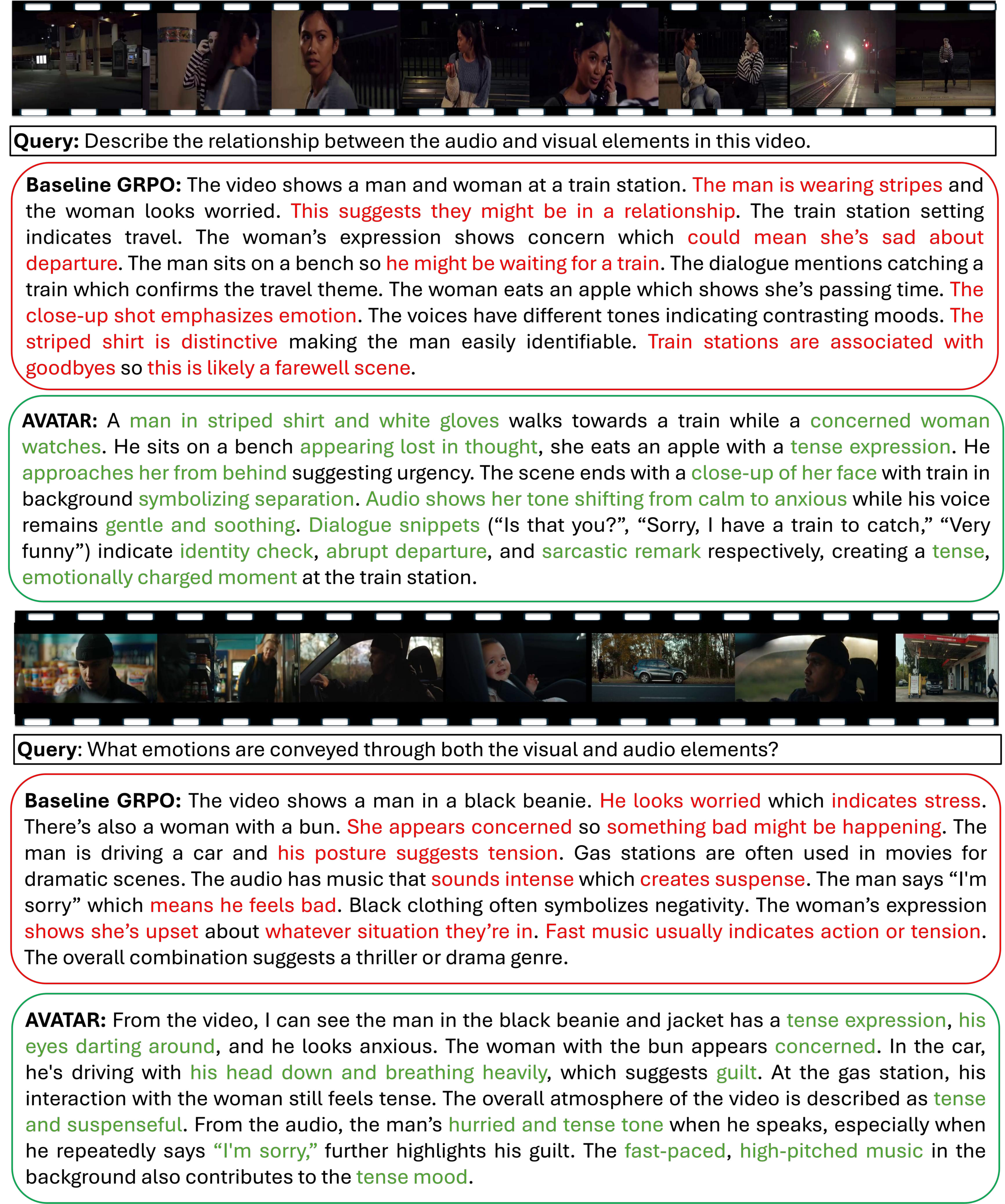}
\caption{\textbf{Qualitative comparison between Baseline GRPO and \modelname{}}. \modelname{} demonstrates better \textbf{cross-modal integration} and \textbf{structured reasoning}, while Baseline GRPO shows \textit{surface-level observations} and \textit{disconnected audio-visual analysis}. Green highlighting shows \modelname{}’s effective synthesis of multimodal cues, red highlighting reveals GRPO’s uniform credit assignment limitations.}
\label{fig:qualitative_analysis}
\end{figure*}

Figure~\ref{fig:qualitative_analysis} illustrates the qualitative differences between baseline GRPO and \modelname{} on audio-visual reasoning tasks. \modelname{} demonstrates \textit{superior cross-modal integration} by linking visual cues (``tense expression, his eyes darting around’’) with audio analysis (``hurried and tense tone when he speaks’’), while baseline GRPO makes \textit{disconnected observations}, such as ``He looks worried which indicates stress’’ without linking modalities. The responses highlight the effectiveness of TAS, evident in the model’s ability to \textit{plan} by establishing comprehensive scene context, focus on relevant cues while avoiding \textit{redundant pattern matching} (e.g., ``Black clothing often symbolizes negativity’’), and \textit{synthesize} complex emotional dynamics rather than relying on \textit{surface-level assumptions} (``something bad might be happening’’). \modelname{} also shows \textit{improved temporal reasoning} by tracking emotional progression (``tone shifting from calm to anxious’’) and \textit{enhanced contextual understanding} through precise dialogue interpretation (``Sorry, I have a train to catch’’ indicating abrupt departure). In contrast, baseline GRPO tends to fall back on \textit{generic genre classifications} (``suggests a thriller or drama genre’’). These qualitative differences validate our framework’s ability to enhance the quality of multimodal reasoning through targeted credit assignment and structured learning signals.

\clearpage
\begin{figure*}[!t]
    \centering
    \begin{tcolorbox}[
        title={Hint Generation Prompt for Hard Samples},
        colframe=black,
        colback=gray!10,
        coltitle=white,
        fonttitle=\bfseries,
        width=\textwidth
    ]
    You are an expert AI tutor. Your task is to provide a concise, strategic hint to a student model that is struggling with a difficult audio-visual reasoning problem. The student model is consistently failing this problem and has stopped exploring new reasoning paths. Your hint should guide it out of a local minimum without solving the problem for it.
    
    Given the following full problem context:
    
    \textbf{[User Query]:} \{The original question posed to the student model.\}
    
    \textbf{[Audio Caption]:} \{A transcript of the relevant audio from the video clip.\}
    
    \textbf{[Video Context]:} \{The full video clip associated with the query.\}
    
    \textbf{[Ground Truth Answer]:} \{The correct final answer to the query.\}
    
    \hspace{2cm}
    
    \textbf{Your Task:}
    Based on the full context and the ground truth, generate a single, short sentence that provides a high-level strategy for how to approach the problem.
    
    \textbf{Constraints:}
    \begin{itemize}
        \item \textbf{DO NOT} reveal or allude to the final answer.
        \item \textbf{DO NOT} perform the reasoning steps yourself.
        \item Your hint must be a strategic suggestion (e.g., ``focus on the audio first,'' ``compare the objects on the left and right,'' ``track the object's movement over time'').
    \end{itemize}
    
    \textbf{Example Hint:}
    \texttt{first locate the object making the sound, then count}
    
    \textbf{Output:}
    Provide ONLY the text of the hint.
    
    \end{tcolorbox}
    \caption{The prompt template used to generate guiding hints for the main policy model.}
    \label{fig:hint-prompt}
\end{figure*}
\begin{figure*}[!t]
    \centering
    \begin{tcolorbox}[
        title={Audio-Visual Localization Judge Prompt},
        colframe=black,
        colback=gray!10,
        coltitle=white,
        fonttitle=\bfseries,
        width=\textwidth
    ]
    You are a meticulous and precise Audio-Visual Grounding Evaluator. Your task is to provide a granular, stepwise evaluation of a model's attempt to localize an object based on audio and visual cues.
    
    Given the following inputs, you will score the model's reasoning and final answer on several criteria.
    
    \textbf{[User Query]:} \{The user's original question to the model.\}\\
    \textbf{[Audio Caption]:} \{A transcript of the relevant audio from the video clip.\}\\
    \textbf{[Ground Truth Answer]:} \{The correct description and location of the target object.\}\\
    \textbf{[Model's Reasoning]:} \{The reasoning process generated by the model inside its \texttt{<think>} tags.\}\\
    \textbf{[Model's Final Answer]:} \{The final location and description from the model's \texttt{<answer>} tag.\}
    
    \hspace{2cm}
    
    \textbf{Your Task:}
    Based on your analysis of all the provided information, you must evaluate the model's performance on the following four criteria. Provide your judgment as a single JSON object with a score from 0.0 (complete failure) to 1.0 (perfect) for each criterion.
    
    \begin{enumerate}
        \item \textbf{Audio Cue Grounding (\texttt{audio\_grounding\_score}):} Did the model's reasoning correctly identify the key descriptive words in the \texttt{[Audio Caption]} that point to the target? (e.g., did it correctly identify ``clapping'' as the key sound?).
        
        \item \textbf{Visual Object Identification (\texttt{visual\_id\_score}):} Based on the audio cue, did the model's reasoning correctly identify the corresponding visual object in the scene? (e.g., after identifying ``clapping,'' did it correctly identify the ``audience'' as the source?).
        
        \item \textbf{Location Accuracy (\texttt{location\_acc\_score}):} How accurate is the final spatial location (e.g., ``Left,'' ``Center'') provided in the \texttt{[Model's Final Answer]} when compared to the \texttt{[Ground Truth Answer]}?
        
        \item \textbf{Caption Correctness (\texttt{caption\_corr\_score}):} How well does the model's final textual description of the object in its \texttt{[Model's Final Answer]} match the \texttt{[Ground Truth Answer]}?
    \end{enumerate}
    
    \textbf{Output Format:}
    Return ONLY a single JSON object with your scores.
    
    \textbf{Example Output:}\\
    \texttt{\{``audio\_grounding\_score'': 0.9, 
    ``visual\_id\_score'': 1.0, 
    ``location\_acc\_score'': 1.0, 
    ``caption\_corr\_score'': 0.8\}}
    
    \end{tcolorbox}
    \caption{The prompt template for our VLM-based judge in Stage $3$. It guides the judge to produce a granular, stepwise reward by scoring the policy model's reasoning on audio grounding, visual identification, location accuracy, and caption correctness.}
    \label{fig:judge_prompt_stage3}
\end{figure*}

\end{document}